\documentclass[11pt]{article}

\usepackage[dvipsnames]{xcolor}

\usepackage{ACL2023}
\usepackage{times}
\usepackage{latexsym}
\usepackage[T1]{fontenc}
\usepackage[utf8]{inputenc}
\usepackage{microtype}
\usepackage{inconsolata}
\usepackage{todonotes}
\usepackage{booktabs}
\usepackage{latexsym}
\usepackage{linguex}
\usepackage{dblfloatfix}
\usepackage{graphicx}
\usepackage{multirow}
\usepackage{caption}
\usepackage{subcaption}
\usepackage[nameinlink]{cleveref}

\newcommand{\aautoref}[1]{\hyperref[#1]{Appendix~\ref*{#1}}}

\usepackage{floatrow}
\newfloatcommand{capbtabbox}{table}[][\FBwidth]

\definecolor{context}{HTML}{E69F00}
\definecolor{verb}{HTML}{D55E00}
\definecolor{options}{HTML}{56B4E9}
\definecolor{mask}{HTML}{009E73}

\title{Causal interventions expose  implicit situation models \\ for commonsense language understanding}

\author{Takateru Yamakoshi$^{1}$, James L. McClelland$^{2}$, Adele E. Goldberg$^{3}$,  Robert D. Hawkins$^{3}$ \\
$^1$The University of Tokyo, $^2$Stanford University, $^3$Princeton University\\
\texttt{jlmcc@stanford.edu}, \texttt{\{takateru,adele,rdhawkins\}@princeton.edu} }

\begin{document}
\maketitle
\begin{abstract}
Accounts of human language processing have long appealed to implicit ``situation models'' that enrich comprehension with relevant but unstated world knowledge.
Here, we apply causal intervention techniques to recent transformer models to analyze performance on the Winograd Schema Challenge (WSC), where a single context cue shifts interpretation of an ambiguous pronoun.
We identify a relatively small circuit of attention heads that are responsible for propagating information from the context word that guides which of the candidate noun phrases the pronoun ultimately attends to.
We then compare how this circuit behaves in a closely matched ``syntactic'' control where the situation model is not strictly necessary.
These analyses suggest distinct pathways through which implicit situation models are constructed to guide pronoun resolution.
\end{abstract}

\section{Introduction}

Language understanding is deeply intertwined with world knowledge.
For example, when reading a sentence like ``the fish ate the worm,'' we can guess that the fish was probably hungrier before eating and that the worm is no longer alive, even though neither property is explicitly mentioned \cite{winograd1972understanding,rumelhart1975notes}. 
Classical psycholinguistic accounts have suggested that such knowledge enters into language understanding through structured schemas called \emph{situation models} \cite{zwaan1998situation,morrow1988interpreting, bransford1971abstraction, schank2013scripts, bower1979scripts, graesser1994constructing,johnson1983mental}
that are dynamically constructed during comprehension.
Put succinctly, a situation models is a representations of the web of entities and relations that are \emph{implied} without being explicitly specified in the literal text \cite{safavi2021relational,piantadosi2022meaning}.

\begin{figure}[t!]
\begin{center}
\includegraphics[width=0.999\linewidth]{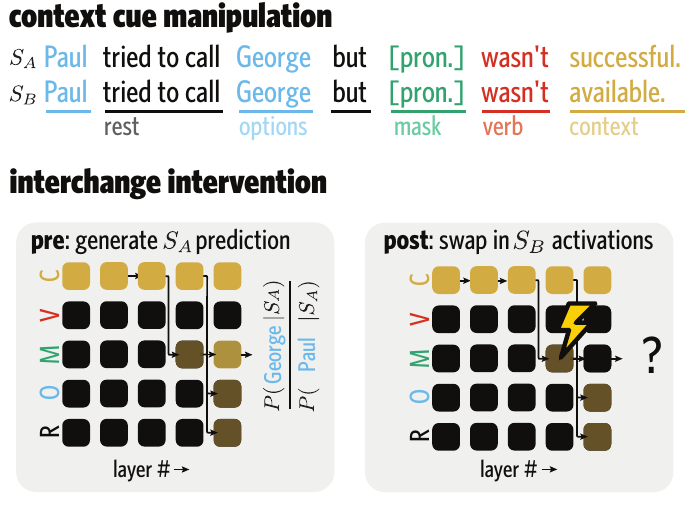}
\caption{We analyze the circuits responsible for performance on Winograd sentences using a series of causal interchange interventions to trace how contextual information is integrated to resolve the masked pronoun.}
\label{fig:front}
\end{center}
\end{figure}

Modern large language models (LLMs) exhibit increasingly impressive performance on ``commonsense'' tasks that seemingly require the use of implicit world knowledge \cite{sap-etal-2019-social,zellers-etal-2019-hellaswag,petroni-etal-2019-language,davison-etal-2019-commonsense,vulic-etal-2020-probing}, yet it is still not clear precisely how that knowledge is accessed and employed. 
Recent interpretability work has explicitly probed and traced individual pieces of world knowledge to highly localized regions of the network \cite[e.g. the birth years of US Presidents;][]{meng2022locating,dai-etal-2022-knowledge}, allowing surgical erasure or editing \cite[e.g.][]{colon2021combining}.
But the kind of world knowledge represented by a situation model is more implicit and, according to classical theories, constructed on the fly for the task at hand. 
Models must somehow determine that a particular relational concept is relevant in the first place \cite{icard2015resource}.

For example, in \autoref{fig:front} the final piece of context (\textit{successful} / \textit{available}) is only able to resolve the pronoun in light of the initial verb phrase ``tried to call''. 
The situation model constructed from an agent Paul \emph{trying} something raises the possible predicates of success or failure, while a patient George \emph{being called} raises the possible predicates of busyness or availability.
Conversely, \emph{Paul}'s availability and \emph{George}'s success are not at issue and therefore not available as interpretations for the pronoun. 
There are, of course, many other attributes that may be invoked from a \emph{tried to call} situation model that are not relevant to resolving the pronoun, involving the kinds of devices used to make calls, the kinds of noises heard when being called, and so on. 
According to this analysis, successfully deriving the appropriate referent in both sentences requires models to \emph{integrate} relevant information across disparate parts of the sentence structure.

In this paper, we conduct an initial exploration of the hypothesis that LLMs have learned to construct implicit situation models. 
As a case study, we conduct a fine-grained analysis of a best-in-class transformer model \cite[ALBERT;][]{Lan2020ALBERT} on a Winograd-like pronoun disambiguation task \cite[][see \autoref{fig:front}]{winograd1972understanding, levesque2011winograd,sakaguchi2021winogrande}.
Winograd sentences are minimal pairs constructed with the property that resolving an ambiguous pronoun requires situational world knowledge outside the scope of the literal text; critically, for our purposes, the pair of sentences differs only at one site, which causes the interpretation to flip.
Controlling for possible confounds, a model that is able to generate sharply different predictions for these pairs can be said to behave as if it has a situation model.\footnote{It is important to distinguish this cautious claim of task-specific functional equivalence from the stronger claim that a neural model constructs the same kind of situation models as humans in general.}
Although the open models that amenable to causal probing do not yet achieve fully human-like performance on this task, their trajectory of increasing functional capabilities raises an important mechanistic question about \emph{how} larger networks achieve these gains. 

Our primary contributions are (1) employing causal probes to identify a sub-circuit of attention heads that are responsible for propagating contextual information, and (2) constructing a set of closely matched controls for Winograd sentences that are resolvable solely using syntactic cues, which we use to validate the specificity of the identified situation model circuit.
Overall, we find some exciting preliminary evidence for meaningfully non-overlapping pathways
while also highlighting the subtleties of probing situation models on a sentence-by-sentence basis. 

\section{Related Work}

\subsection{Implicit world knowledge in LLMs}

A number of recent studies have examined the extent to which neural language models have acquired implicit schemas about the world \cite{li2021implicit,dai-etal-2022-knowledge}, proposed auxiliary tasks to improve coherence \cite{li2022language}, and probed the internal mechanics by which world knowledge influences downstream predictions \cite{meng2022locating,geva2022transformer}.
A smaller set of studies has focused on classical psycholinguistic phenomena: for example, \citet{davis-van-schijndel-2020-discourse} examining relative clause attachment in coreference resolution, and \citet{upadhye2020predicting} examining predictions about which entities are preferred for different verbs.
We approach the problem of implicit situation model representations with a more targeted set of causal intervention techniques, tracing the internal flow of subtle contextual cues in Winograd schemas for the first time. 

\subsection{Probing with causal interventions}

In order to identify interpretable algorithms underlying specific model behaviors, recent studies have employed variants of causal intervention analyses on intermediate representations, targeting syntactic agreement \cite{lakretz2021mechanisms,finlayson-etal-2021-causal,lasri-etal-2022-probing}, relative clause processing \cite{ravfogel-etal-2021-counterfactual}, natural language inference  and compositionality \cite{geiger-etal-2020-neural,geiger-2021,soulos-etal-2020-discovering}, gender bias \cite{vig2020investigating}, sub-word representations \cite{huang2022inducing}, and factual knowledge \cite{dai-etal-2022-knowledge,meng2022locating}.
Following \citet{olah2020zoom}, we use the term \emph{circuit} to capture the explanatory construct in such interpretability studies: a computational subgraph of a neural network consisting of a set of units and connections between them that are causally implicated in a behavior \cite{elhage2021mathematical,olsson2022context}. 
For example, \citet{wang2022interpretability} recently argued that a small circuit of attention heads appears to identify indirect objects, and \citet{wu2023interpretability} revealed a circuit for solving simple numerical reasoning problems.
Building on this family of \emph{interchange intervention} techniques \cite{geiger-etal-2020-neural,mueller2022causal,elazar2021amnesic}, we decompose each head into its query, key and value sub-components \cite{mohebbi2023quantifying} to trace the flow of causally important information for solving Winograd sentences.

\section{Approach}

\subsection{Dataset construction}

We began by extracting the subset of the Winograd Schema Challenge (WSC) appearing in SuperGLUE \cite{wang2019superglue,levesque2011winograd}, as well as the larger, crowd-sourced Winogrande \cite{sakaguchi2021winogrande} dataset.
These datasets contain sentence pairs that differ only at a minimal word or phrase that changes the referent of an ambiguous pronoun earlier in the sentence (see \autoref{fig:front}).\footnote{To facilitate comparison across sentences of different lengths, we will refer to the span of text that is manipulated across the two sentences as the ``context", the candidate referents as the ``options", the pronoun as the ``mask" (since it is masked for the prediction task), the verb immediately following the mask as the ``verb", and the remaining tokens as the ``rest''. For all analyses, we conduct single-token interventions and then aggregate effects within that class.}

\ex.
Paul tried to call George but \texttt{<MASK>} wasn't [\textbf{successful} / \textbf{available}].

These sentences are structured such that the disambiguating context only appears near the end, but otherwise have diverse sentence structure (e.g. the context can be any part-of-speech).
We call these pairs the ``context cue'' condition. 

We then modified these pairs to construct three additional conditions for comparison.
In a ``context+syntax cue" condition, we changed the plurality of one of the noun phrase options such that the pronoun can be resolved by relying on the number signaled by the verb without necessarily requiring situational knowledge.

\ex.
They tried to call George but \texttt{<MASK>} [\textbf{weren't} successful] / [\textbf{wasn't} available].

To remove the availability of world knowledge entirely, we masked out the context span in both sentences to form a ``syntax only" condition.

\ex.
They tried to call George but \texttt{<MASK>} [\textbf{weren't} / \textbf{wasn't}] \texttt{<MASK>}.

Finally, we generated a control condition using semantically equivalent \emph{synonyms}; if results in the other conditions truly reflect world knowledge rather than spurious token-specific features, we should not expect to find any effect in this condition.

\ex.
Paul tried to call George but he [\textbf{wasn't accessible}] / [\textbf{wasn't available}].

In total, we constructed 200 unique pairs of sentences, each of which appears within all 4 conditions (see more examples in \autoref{tab:example_sentences}; further details of dataset construction are provided in Appendix A.)

\begin{table*}[t]
    \centering
    \begin{tabular}{l|c|c|c|c|c|c}
        \multirow{2}{*}{Model} & \multicolumn{2}{c|}{context only} & \multicolumn{2}{c|}{syntax only} & \multicolumn{2}{c}{context+syntax}\\
            & strict [\%] & weak [\%] & strict [\%] & weak [\%] & strict [\%] & weak [\%]\\
        \hline\hline
        bert-base-uncased & 11.0 & 59.5 & 81.5 & 99.5 & 83.5 & \textbf{100} \\
        bert-large-cased & 15.5 & 65.0 & 81.5 & \textbf{100} & 80.5 & 99.5 \\
        roberta-base & 8.5 & 59.0 & 69.5 & \textbf{100} & 72.5 & \textbf{100}  \\
        roberta-large & 12.5 & 70.0 & 64.0 & 99.0 & 66.0 & 99.5 \\
        albert-base-v2 & 12.5 & 56.0 & 60.0 & 95.5 & 84.5 & 99.0\\
        albert-large-v2 & 14.5 & 57.0 & 78.0 & 98.0 & 89.0 & 99.5 \\
        albert-xlarge-v2 & 18.5 & 59.5 & \textbf{90.0} & \textbf{100} & \textbf{91.0} & 99.5 \\
        \textbf{albert-xxlarge-v2} & \textbf{31.5} & \textbf{81.5} & 88.5 & \textbf{100} & 88.5 & \textbf{100} \\
        \hline gpt-4 & 51.0 & -- & -- & -- & --& --\\
        gpt-4 + CoT & 67.3 & -- & -- & -- & -- & -- \\       
        \hline
        human & 94.1 & -- & -- & -- & -- & -- \\
    \end{tabular}
    \caption{Zero-shot performance on Winograd items. The strict metric requires the correct referent to be strictly preferred on both sentences in the pair, so chance is 0\% for a model that is not sensitive to context. The weak metric only requires context to shift the prediction in the correct direction across the pair (even if there is an absolute preference for the incorrect referent), so chance is 50\% for a model that is not sensitive to context. We could not compute weak scores for gpt-4 because raw probabilities are not exposed by the API.}
    \label{tab:model_performance}
\end{table*}

\subsection{Interchange interventions}

To interrogate how exactly masked language models achieve context-sensitive predictions from minimal cues, we applied a causal intervention technique \cite{geiger-etal-2020-neural,mueller2022causal} to map the flow of information from the context word to the masked site where the ultimate prediction is made. 
Specifically, we used an \emph{interchange intervention} to swap intermediate representations across the two contexts (see \autoref{fig:front}), and quantified the effect of the intervention on the model's downstream prediction. 
Given a pair of sentences $(s_A, s_B)$ that differ only at the context token, we mask out the pronoun and score the likelihood of each noun phrase ($N_A$, $N_B$) at the masked position.
For simplicity, we denote the noun phrases such that $N_A$ is the correct referent for sentence $s_A$ and $N_B$ is the correct referent for sentence $s_B$.
For example, $P_\theta(N_A|s_A)$ refers to the likelihood assigned to the (correct) referent $N_A$ at the mask position in sentence $s_A$, where $\theta$ represents the model parameters.
In the case of multi-token noun phrases, we masked out all tokens in the phrase and used the average log probability of each token as the score (see Appendix C).

Although multiple metrics have been proposed for capturing the effect of an intervention \cite[e.g.][]{mueller2022causal}, we use the canonical \emph{odds ratio,} the shift in relative preference for the correct option as a result of the intervention.
We first calculate the baseline preference for the correct referent $N_A$ relative to the incorrect referent $N_B$: $$y_{pre} = \frac{P_\theta(N_A | s_A)}{P_\theta(N_B | s_A)}$$ where $P_\theta()$ represents the model's prediction under pre-intervention representations $\theta$. 
We then measure the same preference after the intervention:
$$y_{post} = \frac{P_{\theta^{+z}}(N_A | s_A)}{P_{\theta^{+z}}(N_B | s_A)}$$
where $P_{\theta^{+z}}()$ represents the model's prediction using post-intervention representations $\theta^{+z}$. 
The odds ratio is then:
$E = y_{pre} / y_{post}$ (or, on a logarithmic scale, $\log E = \log y_{pre} - \log y_{post}$. 
Effects are averaged across the two directions of intervention within each  sentence pair. 
Note that $\log y_{base} > 0$ by definition for pairs where the baseline prediction is correct, but depending on whether the intervention decreases or increases the probability of the correct referent, the causal effect can be positive (indicating the site of the intervention was contributing to the correct prediction) or negative (indicating the site of the intervention was contributing to the incorrect prediction). 

The odds ratio has some desirable properties compared to other measures like the absolute difference in differences \cite[e.g.][]{yin2022interpreting}. 
It does not suffer from ceiling or floor effects, and it is a well-understood measure in classical statistics, as deployed in logistic regression. 
However, it is also insensitive to the absolute values of the probabilities going into the ratio, making it potentially vulnerable to noise in the tails (i.e. when both NPs are very infrequent). 
We believe the odds ratio is a preferred metric \emph{a priori} but we hope future work will better elucidate the advantages and disadvantages of different metrics.

\section{Results}
\subsection{Zero-shot performance evaluation}

There are a multiplicity of different ways of evaluating performance on Winograd sentences.
These shifting criteria may be responsible for inflated claims of state-of-the-art performance \cite{liu-etal-2020-precise,kocijan2022defeat,trichelair-etal-2019-reasonable}.
To set the stage for our causal intervention analyses, we conduct our own stricter zero-shot comparison of recent pre-trained models.
Our strict metric requires the correct referent to be assigned higher probability for \emph{both} sentences, that is, the likelihood ratios must satisfy
$$\frac{P_\theta(N_A | s_A)}{P_\theta(N_B | s_A)} >1\,\, \textrm{ and } \,\,\frac{P_\theta(N_B| s_B)}{P_\theta(N_A | s_B)} >1$$
By jointly considering both sentences in the pair, this metric better captures context-sensitivity.
Note that a context-insensitive model that makes the same prediction for both sentences would receive a score of zero on this metric \cite[]{abdou-etal-2020-sensitivity,elazar-etal-2021-back}.
That is, context-sensitivity is required for the interpretation of a pronoun to be systematically reversed, as required to meet the stricter accuracy criteria.
We also consider a more standard but weaker metric that only requires the prediction to shift in the correct direction, even if there is an absolute bias for the incorrect option; chance is 50\% for this metric:
$$\frac{P_\theta(N_A | s_A)}{P_\theta(N_B | s_A)} > \,\,\frac{P_\theta(N_A| s_B)}{P_\theta(N_B | s_B)}$$

\autoref{tab:model_performance} reports the performance of three masked language models (BERT, RoBERTa and ALBERT) at different sizes, along with the auto-regressive GPT-4, and a newly elicited dataset of $N=199$ human participants (see Appendix B for details).
While all models we consider fall well short of human performance, larger and more recent models tend to perform better overall, with the large ALBERT model achieving up to 81.5\% on the weak criterion. 
We also observe, unsurprisingly, that models perform better overall for the syntax cue condition than for the context cue condition and even better when both cues are combined. 
GPT-4 performs better than other models on the strict criterion, reaching 67.3\% with the use of chain-of-thought prompting \cite{wei2022chain} (mean accuracy aggregated across all 400 individual sentences is 82\%, compared to 87\% human accuracy).

Although models like GPT-4 are only available through an API, limiting our ability to explore causal interventions on their internal representations, we may view other models as lying at earlier points on the same scaling trajectory. 
We report interventions on all open models (\autoref{fig:models}), but we place particular focus on the largest ALBERT model, which achieved the highest zero-shot performance of any open-source model we considered. 
First, we are interested in  examining the circuits underlying \emph{successful} performance, and the small number of sentences for which they make the correct predictions limit statistical power (e.g. BERT is only correct for 22 sentences under the strong criterion).
Second, the ALBERT architecture ties the weights of attention heads across all layers, yielding more interpretable head-wise analyses (i.e. it is meaningful to track the same head $k$ across layers).

\subsection{Layer-wise information flow}

\begin{figure*}[th!]
\begin{center}
\includegraphics[width=0.99\linewidth]{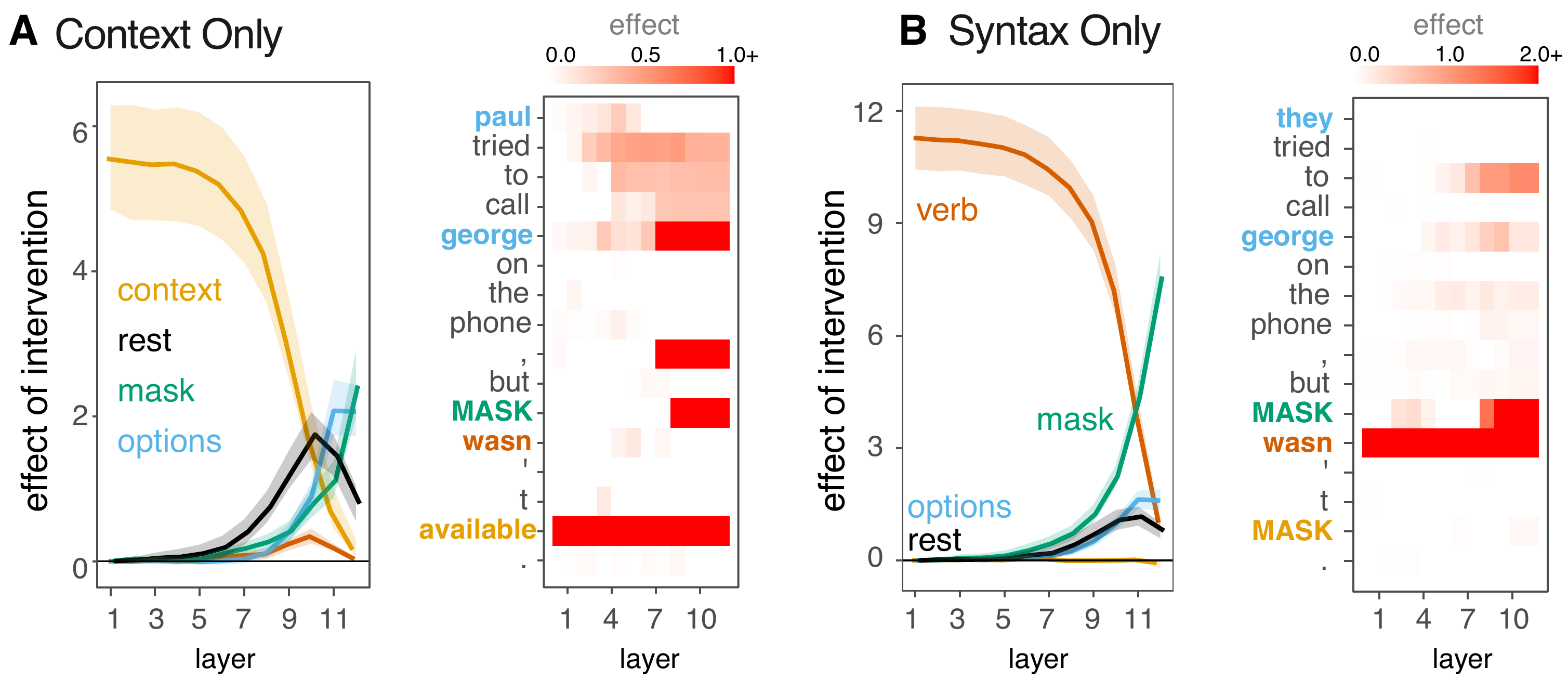}
\caption{The mean effect of causal interchange interventions at each layer and site are shown for (A) the context only condition and (B) the syntax only condition. Error ribbons show bootstrapped 95\% confidence intervals across sentence pairs. Heatmaps show effects of interventions at each individual token for an example sentence.}
\label{fig:layer}
\end{center}
\end{figure*}

To yield insight into failure modes of these models, we would like to develop a more mechanistic understanding of \emph{how} local information from the context word is algorithmically propagated through other sites in the sentence to ultimately arrive at the correct prediction.
In other words, we are interested in probing the transformer \emph{circuits} that allow such minimal context cues to have such large effects on pronoun resolution.
According to classical accounts, situation models must be constructed \emph{dynamically}, as it is not obvious ahead of time which aspects of the situation will be relevant to interpretation. 

\begin{figure}[b!]
\begin{center}
\includegraphics[width=0.99\linewidth]{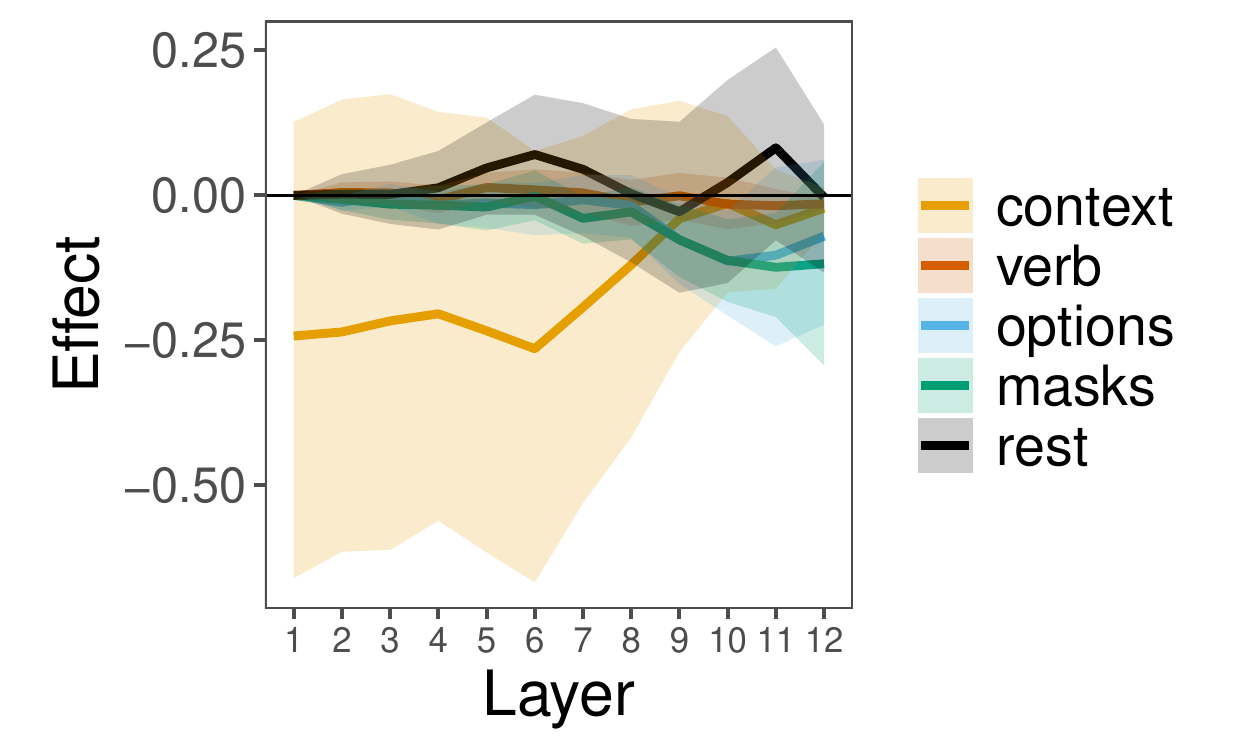}
\vspace{-2em}
\caption{Layerwise intervention for \emph{synonyms} yields no significant effects, indicating that heads are not purely driven by lexical identity. Error ribbons show bootstrapped 95\% confidence intervals across sentence pairs.}
\label{fig:synonym}
\end{center}
\end{figure}

\begin{figure*}[t!]
\begin{center}
\includegraphics[width=0.99\linewidth]{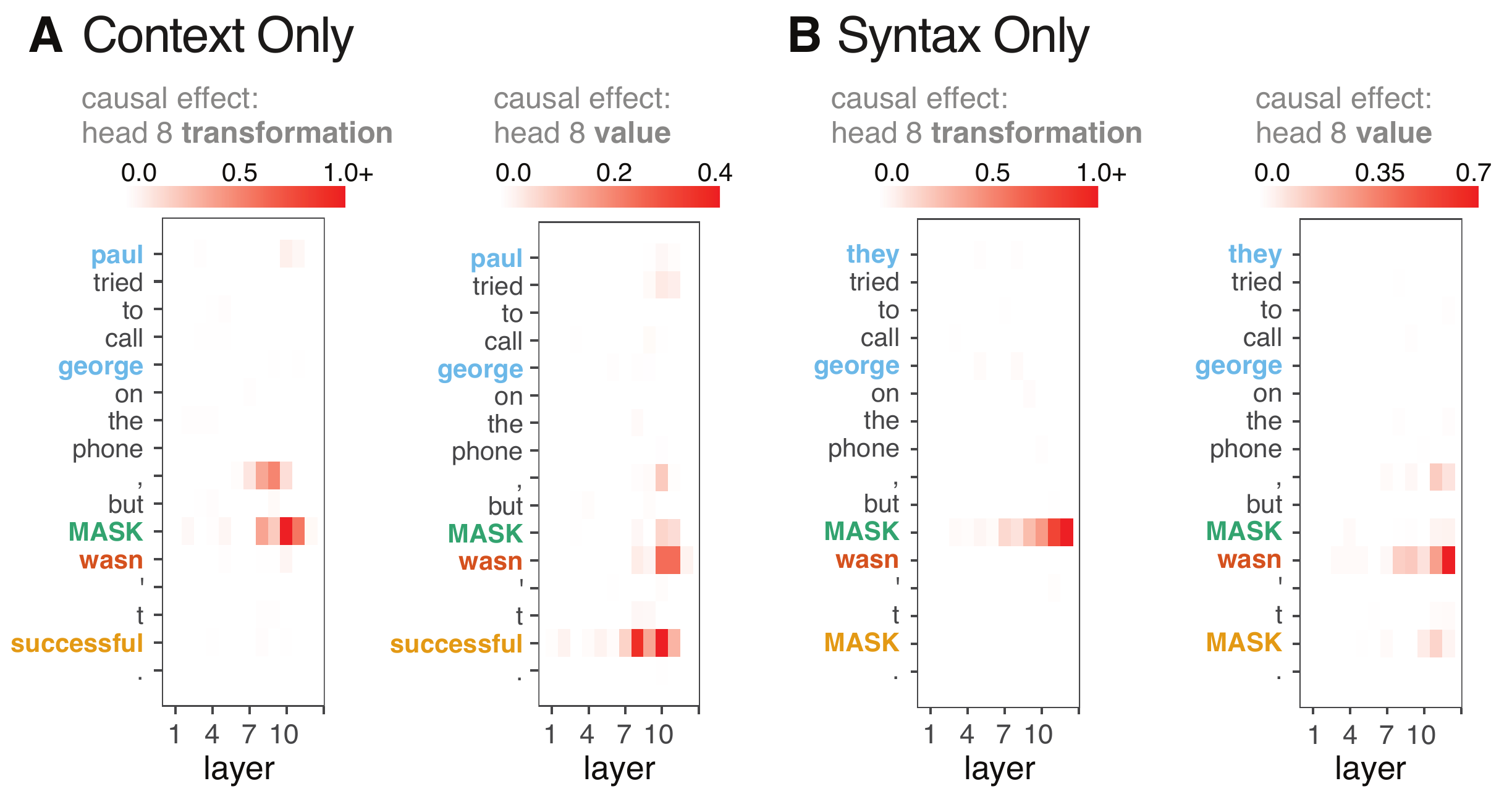}
\caption{Effects of intervening at a single attention head (head 8) for an example sentence in  (A) the context condition and (B) the syntax condition. We intervene at both the point where this head's output is added back to the given token's residual stream (\emph{transformations}) as well as the internal \emph{value} vector this head exports to other tokens when queried. In the syntax condition, this head appears to be primarily responsible for moving information from the verb to the MASK; in the context condition, however, it is implicated at other sites, like the comma token where phrase-level information may be integrated.}
\label{fig:headexamples}
\end{center}
\end{figure*}

We begin by considering the effect of coarse-grained layer-wise intervention.
For each layer and each token, we replaced its vector representation under one sentence with what it would have been at the same layer and token for the other sentence, and measured the extent to which the output prediction changed
(see 
\autoref{fig:layer}). 
Results are shown only for sentences where the model made ``strictly'' correct predictions in both the context condition and the syntax condition; effects are similar for cases where the baseline predictions are ``weakly'' correct (see \autoref{fig:layer_incorrect}).

First, as a sanity check, we observe that intervening at the critical context token at early layers dramatically switches the model's prediction (\autoref{fig:layer}). 
This effect begins to decay around layer 9.
Meanwhile, intervening at the noun phrase options only yielded a significantly non-zero effect on the model's output starting layer 9, $t(57) = 6.1, p<0.05$ (correcting for multiple comparisons), while intervening at the masked pronoun and other positions (rest) yielded effects beginning at layers 6 and 7, respectively ($p<0.05$).
Although effects are localized among ``rest'' tokens in different locations from sentence to sentence, the example in \autoref{fig:layer}A reveals early causal effects in the ``tried to call'' construction and the comma token, which may serve as a neutral site for aggregating phrase-level information\footnote{It has been previously observed that a large amount of attention at late layers is focused on punctuation and other special tokens \citep{kovaleva2019revealing,clark-etal-2019-bert}. Interestingly, though, we found no systematic effects at other annotated elements, such as final periods or \texttt{[CLS]}/\texttt{[SEP]} tokens (\autoref{fig:period}), suggesting that these elements are not consistently involved in the circuit across sentences.}
Finally, as predicted, no significant effects were observed when swapping representations on control pairs with synonymous context words (\autoref{fig:synonym}), indicating that the context-sensitivity observed for Winograd sentences is not purely driven by token-specific sensitivities.
Taken together, these layer-wise effects are consistent with pronoun-relevant information remaining localized in the context until intermediate layers, when it begins to pool in other locations and eventually contextualizes the options to guide attention from the prnoun.

So far, nothing about these comparisons implicates the construction of an implicit situation model --- we may simply be measuring the circuits for pronoun-resolution more generally.
To disentangle these possibilities, we consider our ``syntax only'' control condition, where syntactic agreement information alone is sufficient to make the correct prediction and no contextual information is available.
Unlike the ``context only'' condition, where effects at the rest of the sentence preceded contextualization of the mask and the effect at the options remains stronger than at the mask as late as layer 11 ($t(57) = -5.4, p<0.05$), the ``syntax'' condition is strictly dominated by the effect at the mask starting around layer 7 $(t(57) = 3.8, p <0.05$).
This effect is illustrated in \autoref{fig:layer}B, where we see a much more localized circuit between the auxiliary verb, where a number agreement cue is provided, and the MASK, where the prediction is generated; effects at other sites are much more muted. 
Interestingly, in the combined (context$+$syntax) condition where both cues are available, information at the mask still dominates starting at the same layer as the ``syntax'' condition ($p<0.05$; not shown). 
We found qualitatively similar results for the largest RoBERTa model, but not for the smaller models (see \autoref{fig:models}).

\subsection{Head-wise causal interventions}

These coarse layer-wise analysis suggest that global ``situational'' information may be constructed from the context word, and integrated elsewhere in the sentence, whereas purely syntactic agreement information may be accessed more directly by the masked pronoun.
However, it remains unclear exactly how the transformer accomplishes this task at an algorithmic level, and a lot of complexity was potentially hidden in the ``rest'' category.
In this section, we conduct a finer-grained head-wise analysis of the individual components within each layer that are responsible for routing contextual information between tokens (\autoref{fig:headexamples}).

We consider four internal components: (1) the final \emph{transformation} vector produced by the head that gets concatenated back to other heads before being projected back into the residual stream, (2) the \emph{query} vector that ``imports'' information into a given source site, (3) the \emph{key} vector at other sites which matches the query to yield the attention matrix, and (4) the \emph{value} vector that is ``exported'' from the target site  (see \autoref{fig:schematic} for a schematic).

For ease of interpretability, we will refer to effects of intervening on each of these components in terms of an implied ``source $\rightarrow$ target'' pathway.
For example, if intervening on the \emph{value} exported from the context token has the same effect as intervening on the ultimate \emph{transformation} vector at the options, we loosely say that this is a ``context $\rightarrow$ options'' head.
The effect of swapping the full \emph{transformation} vector is our primary measure of how much pronoun-relevant context-sensitivity is introduced into the residual stream by that head at that layer \emph{overall}, which can then be broken down into its contributing subcomponents.

\begin{figure*}[t!]
\begin{center}
\includegraphics[width=1.05\linewidth]{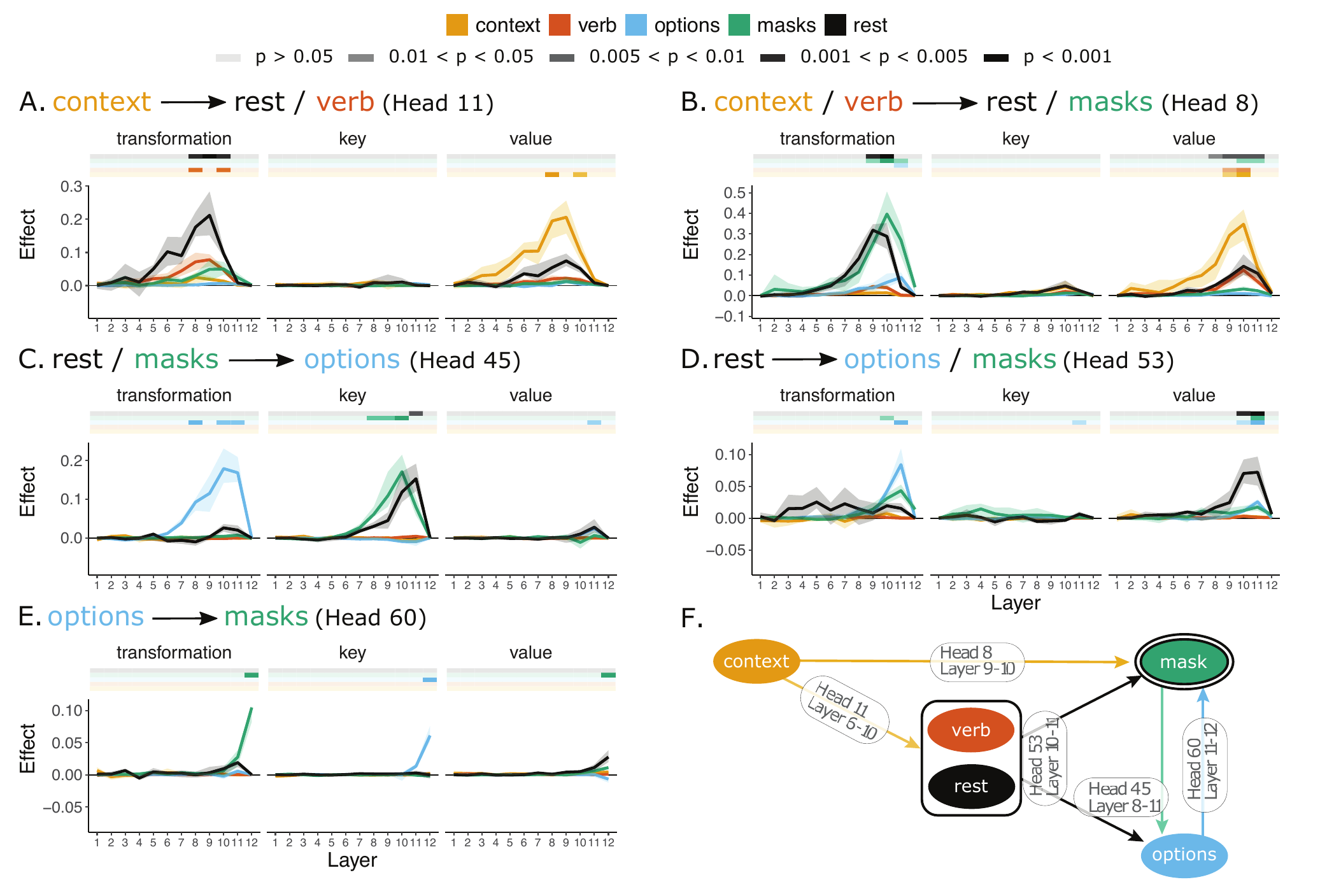}
\caption{Example heads involved in propagation of context information. Error ribbons show bootstrapped 95\% confidence intervals across sentence pairs. Schematic depictions are provided in \autoref{fig:head_context_schematic}.}
\label{fig:head}
\end{center}
\end{figure*}

To build intuition, we first depict the complete profile of causal intervention effects for a single head (head 8) across layers and tokens of an example sentence (\autoref{fig:headexamples}A; see \autoref{fig:roberta} for the same analysis applied to the largest RoBERTa model). 
Consistent with our coarser layer-wise analysis above, we find that intervening on transformations at the comma token around layer 8 (and a few layers later at the MASK token) significantly interferes with the model's ability to correctly resolve the referent. 
We then gained further insight through targeted interventions on the sub-components used to calculate the transformation vector. 
Specifically, we find that intervening on the \emph{value} vector exposed at the context word (\emph{successful}) accounts for a significant proportion of the total effect at middle layers. 
This effect appears to be largely restricted to the context-only condition; causal interventions on the same head in the syntax-only condition (\autoref{fig:headexamples}B) reveal only the more local ``verb $\rightarrow$ mask'' path.

Moving to a more systematic analysis, we find that a set of 23 heads (less than half of the 64 total heads) show a significant effect of intervening on the \emph{transformation} for least one layer and one site ($p<0.05$, after correcting for multiple comparisons;  15 survive at the $p<0.01$ level; 13 at the $p<0.005$ level; and 11 at the $p<0.001$ level).
Examining the internal components of each of these attention heads (keys, queries, values) allows us to construct a preliminary computational graph of how contextual cues eventually propagate to the masked site, aggregating over many sentences. 
\autoref{fig:head} shows the layer-wise profile for five representative heads, each representing a systematic link between a particular set of sites (see \autoref{fig:zrep_all_context} for all heads).
Roughly, this graph suggests that the model begins by shifting information from the context site into ``neutral'' sites throughout the rest of the sentence via head 8 (e.g. value vector at context word, layer 10 : $t(57)=6.1, p<0.05$) and head 11 (value vector at context word, layer 6: $t(57) = 4.9, p<0.05$).
Shortly thereafter, head 45  exports that information to the two noun phrases (options) that are the possible referents of the pronoun (transformation vector at options, layer 8 : $t(57)=6.9,p<0.05$) and head 53 (transformation vector at options, layer 11 , $t(57) = 7.1, p<0.05$). 
Then, at the final layer of the model, head 60 preferentially attends from the mask to one of the two options, determining the ultimate prediction (key vector at layer 11 options, $t(57) = 8.0, p<0.05$).

\begin{figure}[t!]
\begin{center}
\includegraphics[width=0.99\linewidth]{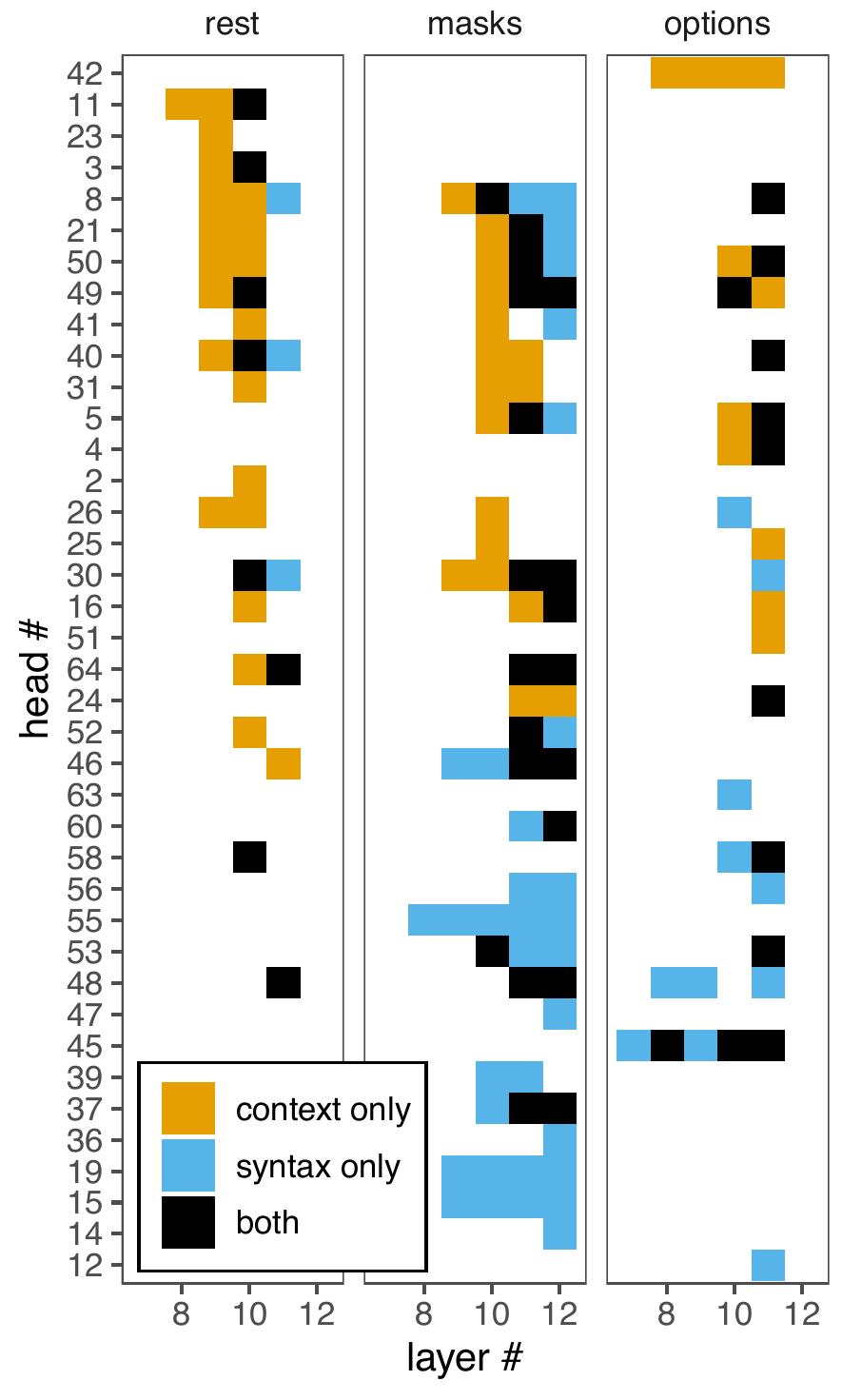}
\vspace{-2em}
\caption{Specificity of head-wise effects across layers and sites. Orange cells only yielded causal effects for the context condition, blue cells only for the syntax condition, and black cells for both conditions. All colored cells are  significantly different from zero at Bonferroni-corrected $p<0.005$ level. Heads are ordered by the earliest layer at which context-specificity appears.}
\label{fig:specificity}
\end{center}
\end{figure}

\subsection{Analysis of context-specificity}

Critically, although some of these heads are also implicated in the syntax-only condition (see \autoref{fig:zrep_all_verb} for full profiles), there are dramatic quantitative differences in the pattern of these heads across layers.
The specificity profile is shown for all active heads in \autoref{fig:specificity}, where specificity is defined with respect to the context vs. syntax comparison: an orange cell indicates `context-specificity' (i.e. selective activation only in the context condition) while a blue cell indicates `syntax-specificity' (i.e. selective activation only in the syntax condition).
Broadly, we observe that many heads, including those highlighted above, yield significant context-specific effects at \emph{earlier layers}, especially at ``rest'' sites (left panel; see \autoref{fig:roberta_all} for an analysis of RoBERTa). 
Conversely, there are a number of heads that are yield syntax-specific effects at ``mask'' sites (middle panel).
Although this pattern of specificity remains highly exploratory, it suggests that the network may have learned different pathways for pronoun resolution: cues like verb number information are directly exposed to guide the mask token, while contextual information must be integrated with other relevant semantic cues from elsewhere in the sentence.
In other words, the latter may require constructing and querying a rudimentary situation model.

\section{Discussion}

In this paper, we presented a preliminary investigation of the transformer circuits underlying performance on Winograd sentences, where minimal contextual cues must be used to resolve an ambiguous pronoun. 
We applied fine-grained causal interventions to identify a circuit of attention heads that are responsible for propagating information from the context cue to the possible referents, which appears to be at least partially distinct from the circuit used to propagate agreement cues in our closely matched syntax-only baseline. 

It still remains to be seen whether the circuits we have identified should be interpreted as evidence for a \emph{bona fide} situation model. 
First, it is possible that even on carefully debiased sets of examples, models like ALBERT are still relying on lexical shortcuts.
For example, presented with ``MASK was tasty'' we would prefer the referent \emph{pie} over the referent \emph{boat}, knowing nothing about the rest of the sentence. 
If so, we may be measuring the circuit for those lexical preferences rather than for anything like a situation model (but see Appendix D for an additional analysis suggesting that this kind of phenomenon is unlikely to be driving the observed effects).

Second, situation models present a well-known example of the frame problem \cite{mccarthy1969some,pylyshyn1987robot} --- it is impossible to explicitly enumerate every property in the world and how it is (or is not) related to every other. 
Hence, any model of interpretation must be ``lazy'' in some sense, only introducing relations that are \emph{relevant} for the task at hand (e.g. the concept of availability for the ``Paul tried to call George'' example). 
When the task at hand is simply pronoun resolution, it is possible that an extremely minimal situation model may be sufficient; in the longer term, it will be important to explore settings that require richer or less accessible interpretations.
Finally, although we focused on implicit situation models constructed internally, it may also be possible to expose situation models more explicitly in larger auto-regressive models through chain-of-thought prompting \cite{talmor2020leap} which do not require model internals to be public.

\section*{Limitations}

Although our single-site interchange interventions provide causal evidence that particular sub-circuits are necessary for a particular downstream behavior, this technique has known limitations addressed by recent Distributed Alignment Search (DAS) approaches \cite{geiger2023finding}. 
First, it will over-count certain ``synergies:'' when a single effect is jointly produced by the conjunction of multiple heads acting in concert, we will identify all heads as making distinct contributions to the circuit.
Second, it will under-count ``redundancies:'' if there are multiple heads that are individual sufficient to produce the effect, then no single head will be detected as strictly necessary. 
Ideally, rather than single-site interventions, we would explore all combinations of different heads to find minimal spanning sets that are both necessary and sufficient, but this procedure becomes intractable given the number of heads, requiring more sophisticated optimization-based approaches to find promising sets \cite[e.g.][]{2021are,de2021sparse}.

\section*{Ethics Statement}

All existing datasets (SuperGLUE and Winogrande) and models (BERT, RoBERTa, and ALBERT) were employed according to their intended research focus, and our targeted probing dataset is intended to be used for similar purposes in future work.
Because it was constructed by a combination of automatic and manual processes by the authors, it contains no additional information that could uniquely identify any individuals. %, to evaluate language models on their ability to build and use situation models.
%Our constructed dataset is intended to be used for similar purposes.
More broadly, we employ causal interventions to evaluate how models perform a challenging commonsense reasoning task, with the aim of building stronger links to classic work in cognitive science and psycholinguistics.
However, these causal interventions may pose some risk if used adversarially to tamper with public models or expose private information.
Further, the WSC and Winogrande datasets we use to probe situation models have been constructed within specific cultural settings (e.g. by NLP researchers and largely US-based crowd-workers) and are not intended to be universal or representative of situational competency: a wider diversity of culturally-specific stimuli is needed.

\section*{Acknowledgements}

We thank Yohei Oseki, Ryo Yoshida, Felix Hill, Devon Wood-Thomas, Aalok Sathe, Claire Bergey, and the Language Computational Cognitive Science Laboratory at UTokyo for thoughtful discussions and comments.
This work was supported by the Princeton Data Driven Social Science initiative, a Seed Grant from the Princeton-UTokyo Strategic Partnership and a C.V. Starr Fellowship to RDH.
Code \& data are available at \url{https://github.com/taka-yamakoshi/situation-models}. 
An interactive demonstration of these causal interventions is available at: \url{https://huggingface.co/spaces/taka-yamakoshi/causal-intervention-demo}.

\bibliography{acl2023}
\bibliographystyle{acl_natbib}

\renewcommand{\thefigure}{S\arabic{figure}}
\renewcommand{\thetable}{S\arabic{table}}
\setcounter{table}{0}
\setcounter{figure}{0}

\section*{Appendix A: Dataset Construction}
\label{app:dataset}

We began with 70 sentences from the WSC subset of SuperGLUE and 9248 sentences from the \texttt{debiased} portion of the Winogrande train split (see Appendix D for discussion of debiasing procedures).
Many of the sentences from Winogrande were singletons, as the other sentence in the pair had been removed in the debiasing procedure, leaving 1361 intact pairs.
We then excluded sentences where we could not automatically obtain the index of the context word or options in the tokenized sequence (many of them had subtle differences in the sentence pair aside from the context word, with respect to the choice of words/punctuation, which are likely not intended by the original annotators of Winogrande), leaving 1140 pairs.

In order to create a closely controlled ``syntax" cue condition, we took the subset where the verb after the pronoun is either an auxiliary (``to be'' or ``to have'') or in present tense, indicating whether the masked pronoun is singular or plural.
These verbs were identified using spaCy (version 3.2.1) \cite{honnibal2020spacy}.
We also ensured at this stage that all sentences had the same NP1-NP2-mask-context-verb ordering.
This step reduced the number of sentence pairs to 38 and 167 for SuperGLUE and Wingorande respectively.

The NPs and the verbs of these extracted sentences were then manually modified to create the agreement cue in the ``syntax'' condition, while the context words were manually modified to create the synonym condition.
We excluded 5 out of 167 sentence pairs at this step, as we were unable to change the plurality of the NPs while preserving the semantic meaningfulness of the sentence (e.g. ``James can count all the numbers on his fingers because the [MASK] are few/many.'').
Finally, because the interchange intervention requires both sentences in a given pair to have exactly the same overall number of WordPiece tokens (in order to independently swap query, key or value features), we lightly modified the contexts to satisfy this condition.
This step did not reduce the number of sentence pairs.

The resulting dataset contains 200 sentence pairs each for the context, syntax and context+syntax cue conditions (38 from SuperGLUE and 162 from Winogrande) and 400 sentence pairs for the synonym condition (as we generated a synonym for each individual sentence in the pair).
This dataset is available on Github (under MIT License)\footnote{\url{https://github.com/taka-yamakoshi/situation-models}}.

\section*{Appendix B: Zero-shot evaluation details}

We used Hugging Face \texttt{transformers}\footnote{Apache License 2.0} (version 4.16.2) \cite{wolf-etal-2020-transformers} for all masked language models.
One of the key features of ALBERT \cite{Lan2020ALBERT}, unlike the original BERT \cite{devlin2018bert} or the subsequent RoBERTa \cite{liu2019roberta}, is that it shares model parameters across layers, which makes it possible to meaningfully compare the same head in different layers.
Relative to the next-largest ALBERT model, the \texttt{xxlarge} model we consider allocates parameters toward increased ``width'' rather than ``depth,'' with larger hidden state dimension (4096 vs. 2048) and more attention heads (64 vs. 16) packed into a smaller number of layers (12 vs. 24).
Running the entire analysis took ~200 hours using 2 GPUs (NVIDIA TITAN X (Pascal)).

For a more state-of-the-art reference point, we also examined the performance of the most recent OpenAI GPT-4 model (using the provided snapshot from March 14, 2023). 
Because GPT-4 is auto-regressive and the disambiguating context in Winograd sentences typically follows the pronoun, we create a prompted variant of the text as suggested by the original proposers of the Winograd challenge \cite{levesque2011winograd}. 
That is, we presented the full sentence along with a question: ``The trombone did not fit in the suitcase because it was too large. What was too large, the trombone or the suitcase?'' 
and coded the model's free-text responses for matching either of the NP options (or ``other'' if it did not match either option). 
We used a temperature of 1 and generated 50 responses for each question to account for sampling variation. 
For our chain-of-thought experiment, we used the following prompt containing a single example:

\begin{quote}
\small
\texttt{
You are going to be shown a sentence and asked to fill in a blank (with an explanation). For example, consider the sentence, "The man tried to put the tuba in a suitcase but <blank> was too small." We will ask whether the <blank> should be interpreted as the TUBA or the SUITCASE.
The <blank> will be literally ambiguous, but you can use knowledge about the world to figure it out. In this case, we can reason as follows. The suitcase is a container, the man is attempting to fit a tuba in that container, and bigger things don't fit when containers are too small. In addition, tubas are generally bigger than suitcases. Therefore the answer is SUITCASE. What about the following sentence?}
\end{quote}

Finally, because our specific set of Winograd items had not previously been benchmarked against human performance, we elicited free-response judgments from a $N=199$ human participants.
We matched the wording of instructions closely to the prompt we used with GPT-4, and we coded responses in the same way.
Each sentence received approximately 20 judgments, we ensured no participant saw more than one sentence from any given Winograd pair, and the order of presentation of the options was counterbalanced (i.e. half of participants saw ''the trombone or the suitcase?'' while the other half saw ''the suitcase or the trombone?'').

\section*{Appendix C: Handling multi-token NPs}
In order to evaluate the model's ability to perform pronoun resolution, we replaced the pronoun with the mask token and compared the log probability of each NP option at the masked site.
While this was straightforward in most cases, 28 of the 200 examples contained a multi-token option.
We handled these cases by introducing multiple masks at the pronoun.
We then calculated an average token-wise score probability in the following way.
Suppose the option noun phrase $NP$ consists of 3 tokens ($w_k$, $w_{k+1}$ and $w_{k+2}$) starting at the $k$th site ($s_k$, $s_{k+1}$ and $s_{k+2}$), where the pronoun is at the $l$th site ($s_l$).
We then replaced the pronoun with three mask tokens and calculated the average log probability as\\ \\
$\frac{1}{3}\small{[\log P(s_l=w_k \,\,\,\,\, | s_{l+1}, s_{l+2}=\texttt{mask}, s_{-l}=w_{-l})}\\
\small{+\log P(s_{l+1}=w_{k+1} | s_{l}, s_{l+2}=\texttt{mask}, s_{-l}=w_{-l})}\\
\small{+\log P(s_{l+2}=w_{k+2} | s_{l}, s_{l+1}=\texttt{mask}, s_{-l}=w_{-l})
]}$,\\ \\
where $s_{-l}=w_{-l}$ denotes that all sites other than $s_{l}$ are filled with the corresponding tokens in the original sentence.
Taking the average log probability effectively controls for the additional number of tokens. 
One concern with this method is that, if part of the noun phrase is a functional word (e.g. a determiner like \emph{the}), it may drag down the observed effect of the entire phrase.
While this is not ideal, we confirmed our results are not sensitive to other ways of aggregating across tokens and believe that aggregating across multiple tokens in this way is a conservative choice relative to other methods, like taking the maximum effect across each of the component tokens. 

\section*{Appendix D: Checking dataset biases}

A recurring concern in the literature on Winograd tasks is that many examples can be fully resolved using lexical ``shortcuts'' that do not require implicit world knowledge. 
We took several steps to ensure that our results are not explained by these biases.
First, we used the debiased train split of Winogrande, which used an approach known as {\sc AfLite} to filter out examples that could be solved using only the isolated embedding of the context word \cite{sakaguchi2021winogrande}.
Second, due to differences in our mask-prediction task and the fine-tuning regime used by \citeauthor{sakaguchi2021winogrande}, we conducted our own analysis of bias in our smaller dataset. 

We devised a simple alternative approach to test whether the isolated context embeddings (token representations at layer 0) had lexical preferences for the options.
We first calculated  correlations (and Euclidean distances) between the embedding vectors at the context and each of the options.
If the context is more similar to $N_1$ (i.e. larger correlation or smaller distance), we say $N_1$ is predicted as the referent, while if the context is more similar to $N_2$, we say it predicts $N_2$.
This simple method made correct predictions in 15.5 \% (using correlation) or 16.5\% (using Euclidean distance) of our 200 sentence pairs.
As expected, this performance is significantly worse than the output MASK prediction.
More importantly, out of the 63 sentence pairs we consider where correct predictions were made by the contextualized model, only 9 (10) were resolvable using word embeddings, based on correlation (distance).
This analysis suggests that while some lexical shortcuts may still exist in our data, these biases are likely not driving our results.

\begin{table*}[]
        \small
    \centering
    \begin{tabular}{c|l}
        & sentence pair\\
        \hline \hline
        \multirow{2}{*}{A} & 
        I poured water from the \textcolor{options}{\textbf{bottle}} into the \textcolor{options}{\textbf{cup}} until the \textcolor{mask}{$<$\textbf{mask}$>$} \textcolor{verb}{\textbf{was}} \textcolor{context}{\textbf{empty}}.\\
        & 
        I poured water from the \textcolor{options}{\textbf{bottle}} into the \textcolor{options}{\textbf{cup}} until the \textcolor{mask}{$<$\textbf{mask}$>$} \textcolor{verb}{\textbf{was}} \textcolor{context}{\textbf{full}}.\\
        \hline
        \multirow{2}{*}{B} & 
        I poured water from the \textcolor{options}{\textbf{bottle}} into the \textcolor{options}{\textbf{cups}} until the \textcolor{mask}{$<$\textbf{mask}$>$} \textcolor{verb}{\textbf{was}} \textcolor{context}{\textbf{empty}}.\\
        & I poured water from the \textcolor{options}{\textbf{bottle}} into the \textcolor{options}{\textbf{cups}} until the \textcolor{mask}{$<$\textbf{mask}$>$} \textcolor{verb}{\textbf{were}} \textcolor{context}{\textbf{empty}}.\\
        \hline
        \multirow{2}{*}{C} & 
        I poured water from the \textcolor{options}{\textbf{bottle}} into the \textcolor{options}{\textbf{cups}} until the \textcolor{mask}{$<$\textbf{mask}$>$} \textcolor{verb}{\textbf{was}} \textcolor{context}{$<$\textbf{mask}$>$}.\\
        &
        I poured water from the \textcolor{options}{\textbf{bottle}} into the \textcolor{options}{\textbf{cups}} until the \textcolor{mask}{$<$\textbf{mask}$>$} \textcolor{verb}{\textbf{were}} \textcolor{context}{$<$\textbf{mask}$>$}.\\
        \hline
        \multirow{2}{*}{$D_1$} & 
        I poured water from the \textcolor{options}{\textbf{bottle}} into the \textcolor{options}{\textbf{cup}} until the \textcolor{mask}{$<$\textbf{mask}$>$} \textcolor{verb}{\textbf{was}} \textcolor{context}{\textbf{empty}}.\\
        & 
        I poured water from the \textcolor{options}{\textbf{bottle}} into the \textcolor{options}{\textbf{cup}} until the \textcolor{mask}{$<$\textbf{mask}$>$} \textcolor{verb}{\textbf{was}} \textcolor{context}{\textbf{emptied}}.\\
        \hline
        \multirow{2}{*}{$D_2$} & 
        I poured water from the \textcolor{options}{\textbf{bottle}} into the \textcolor{options}{\textbf{cup}} until the \textcolor{mask}{$<$\textbf{mask}$>$} \textcolor{verb}{\textbf{was}} \textcolor{context}{\textbf{filled}}.\\
        & 
        I poured water from the \textcolor{options}{\textbf{bottle}} into the \textcolor{options}{\textbf{cup}} until the \textcolor{mask}{$<$\textbf{mask}$>$} \textcolor{verb}{\textbf{was}} \textcolor{context}{\textbf{full}}.\\
        \hline \hline
        
        \multirow{2}{*}{A} &
        \textcolor{options}{\textbf{Joe}} paid \textcolor{options}{\textbf{the detective}} because \textcolor{mask}{$<$\textbf{mask}$>$} \textcolor{verb}{\textbf{has}} \textcolor{context}{\textbf{received}} the final report on the case.\\
        & 
        \textcolor{options}{\textbf{Joe}} paid \textcolor{options}{\textbf{the detective}} because \textcolor{mask}{$<$\textbf{mask}$>$} \textcolor{verb}{\textbf{has}} \textcolor{context}{\textbf{delivered}} the final report on the case.\\
        \hline
        \multirow{2}{*}{B} & 
        \textcolor{options}{\textbf{Joe}} paid \textcolor{options}{\textbf{the detectives}} because \textcolor{mask}{$<$\textbf{mask}$>$} \textcolor{verb}{\textbf{has}} \textcolor{context}{\textbf{received}} the final report on the case.\\
        & 
        \textcolor{options}{\textbf{Joe}} paid \textcolor{options}{\textbf{the detectives}} because \textcolor{mask}{$<$\textbf{mask}$>$} \textcolor{verb}{\textbf{have}} \textcolor{context}{\textbf{delivered}} the final report on the case.\\
        \hline
        \multirow{2}{*}{C} & 
        \textcolor{options}{\textbf{Joe}} paid \textcolor{options}{\textbf{the detectives}} because \textcolor{mask}{$<$\textbf{mask}$>$} \textcolor{verb}{\textbf{has}} \textcolor{context}{$<$\textbf{mask}$>$} the final report on the case.\\
        & 
        \textcolor{options}{\textbf{Joe}} paid \textcolor{options}{\textbf{the detectives}} because \textcolor{mask}{$<$\textbf{mask}$>$} \textcolor{verb}{\textbf{have}} \textcolor{context}{$<$\textbf{mask}$>$} the final report on the case.\\
        \hline
        \multirow{2}{*}{$D_1$} & 
        \textcolor{options}{\textbf{Joe}} paid \textcolor{options}{\textbf{the detective}} because \textcolor{mask}{$<$\textbf{mask}$>$} \textcolor{verb}{\textbf{has}} \textcolor{context}{\textbf{received}} the final report on the case.\\
        & 
        \textcolor{options}{\textbf{Joe}} paid \textcolor{options}{\textbf{the detective}} because \textcolor{mask}{$<$\textbf{mask}$>$} \textcolor{verb}{\textbf{has}} \textcolor{context}{\textbf{obtained}} the final report on the case.\\
        \hline
        \multirow{2}{*}{$D_2$} & 
        \textcolor{options}{\textbf{Joe}} paid \textcolor{options}{\textbf{the detective}} because \textcolor{mask}{$<$\textbf{mask}$>$} \textcolor{verb}{\textbf{has}} \textcolor{context}{\textbf{provided}} the final report on the case.\\
        & 
        \textcolor{options}{\textbf{Joe}} paid \textcolor{options}{\textbf{the detective}} because \textcolor{mask}{$<$\textbf{mask}$>$} \textcolor{verb}{\textbf{has}} \textcolor{context}{\textbf{delivered}} the final report on the case.\\
        \hline \hline

        \multirow{2}{*}{A} &
        As \textcolor{options}{\textbf{Andrea}} in the crop duster passed over \textcolor{options}{\textbf{Susan}}, \textcolor{mask}{$<$\textbf{mask}$>$} \textcolor{verb}{\textbf{was}} able to spot the landing \textcolor{context}{\textbf{strip}}.\\
        &
        As \textcolor{options}{\textbf{Andrea}} in the crop duster passed over \textcolor{options}{\textbf{Susan}}, \textcolor{mask}{$<$\textbf{mask}$>$} \textcolor{verb}{\textbf{was}} able to spot the landing \textcolor{context}{\textbf{gear}}.\\
        \hline
        \multirow{2}{*}{B} &
        As \textcolor{options}{\textbf{Andrea}} in the crop duster passed over \textcolor{options}{\textbf{her parents}}, \textcolor{mask}{$<$\textbf{mask}$>$} \textcolor{verb}{\textbf{was}} able to spot the landing \textcolor{context}{\textbf{strip}}.\\
        &
        As \textcolor{options}{\textbf{Andrea}} in the crop duster passed over \textcolor{options}{\textbf{her parents}}, \textcolor{mask}{$<$\textbf{mask}$>$} \textcolor{verb}{\textbf{were}} able to spot the landing \textcolor{context}{\textbf{gear}}.\\
        \hline
        \multirow{2}{*}{C} &
        As \textcolor{options}{\textbf{Andrea}} in the crop duster passed over \textcolor{options}{\textbf{her parents}}, \textcolor{mask}{$<$\textbf{mask}$>$} \textcolor{verb}{\textbf{was}} able to spot the landing \textcolor{context}{$<$\textbf{mask}$>$}.\\
        &
        As \textcolor{options}{\textbf{Andrea}} in the crop duster passed over \textcolor{options}{\textbf{her parents}}, \textcolor{mask}{$<$\textbf{mask}$>$} \textcolor{verb}{\textbf{were}} able to spot the landing \textcolor{context}{$<$\textbf{mask}$>$}.\\
        \hline
        \multirow{2}{*}{$D_1$} &
        As \textcolor{options}{\textbf{Andrea}} in the crop duster passed over \textcolor{options}{\textbf{Susan}}, \textcolor{mask}{$<$\textbf{mask}$>$} \textcolor{verb}{\textbf{was}} able to spot the landing \textcolor{context}{\textbf{strip}}.\\
        &
        As \textcolor{options}{\textbf{Andrea}} in the crop duster passed over \textcolor{options}{\textbf{Susan}}, \textcolor{mask}{$<$\textbf{mask}$>$} \textcolor{verb}{\textbf{was}} able to spot the landing \textcolor{context}{\textbf{field}}.\\
        \hline
        \multirow{2}{*}{$D_2$} &
        As \textcolor{options}{\textbf{Andrea}} in the crop duster passed over \textcolor{options}{\textbf{Susan}}, \textcolor{mask}{$<$\textbf{mask}$>$} \textcolor{verb}{\textbf{was}} able to spot the landing \textcolor{context}{\textbf{wheels}}.\\
        &
        As \textcolor{options}{\textbf{Andrea}} in the crop duster passed over \textcolor{options}{\textbf{Susan}}, \textcolor{mask}{$<$\textbf{mask}$>$} \textcolor{verb}{\textbf{was}} able to spot the landing \textcolor{context}{\textbf{gear}}.\\
        
    \end{tabular}
    \caption{Examples sentence pairs in each condition (A: context / B: context+syntax / C: syntax / $D_1$: synonym 1 / $D_2$: synonym 2). For synonym 1, we changed the context of sentence 2 to match that of sentence 1 and for synonym 2, we changed the context of sentence 1 to match that of sentence 2. We report the aggregated results of both synonym conditions in \autoref{fig:synonym}}
    \label{tab:example_sentences}
\end{table*}

\begin{figure*}[th!]
\begin{center}
\includegraphics[width=0.99\linewidth]{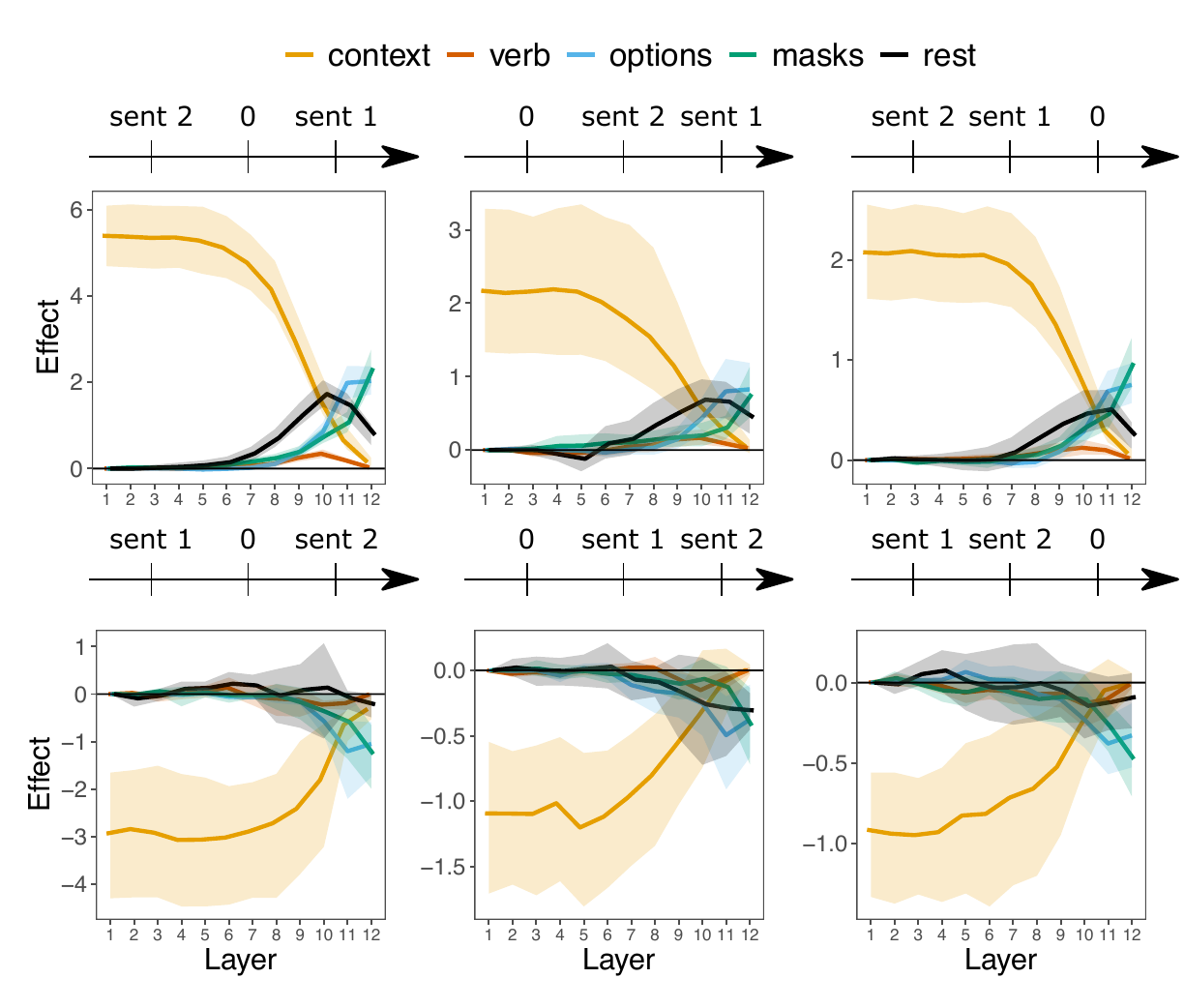}
\caption{Effects of swapping each layer for cases where the model predicted incorrectly. The line at the top shows the sign of the log likelihood ratio $\textrm{log}(L_{\theta}(N_A | s)/L_{\theta}(N_B | s))$ in each sentence $s$.  The top left panel shows the cases where the model predicted correctly for both sentences (positive for $s_A$ and negative for $s_B$).
The cases where the model made incorrect prediction can be categorized into five different groups, depending on the signs of the log likelihood ratios. The sign of the effect is flipped for three panels in the bottom row reflecting the fact that the log likelihood ratio was larger for sentence 2. Error ribbons show bootstrapped 95\% confidence intervals across sentence pairs.}
\label{fig:layer_incorrect}
\end{center}
\end{figure*}

\begin{figure*}[th!]
\begin{center}
\includegraphics[width=0.99\linewidth]{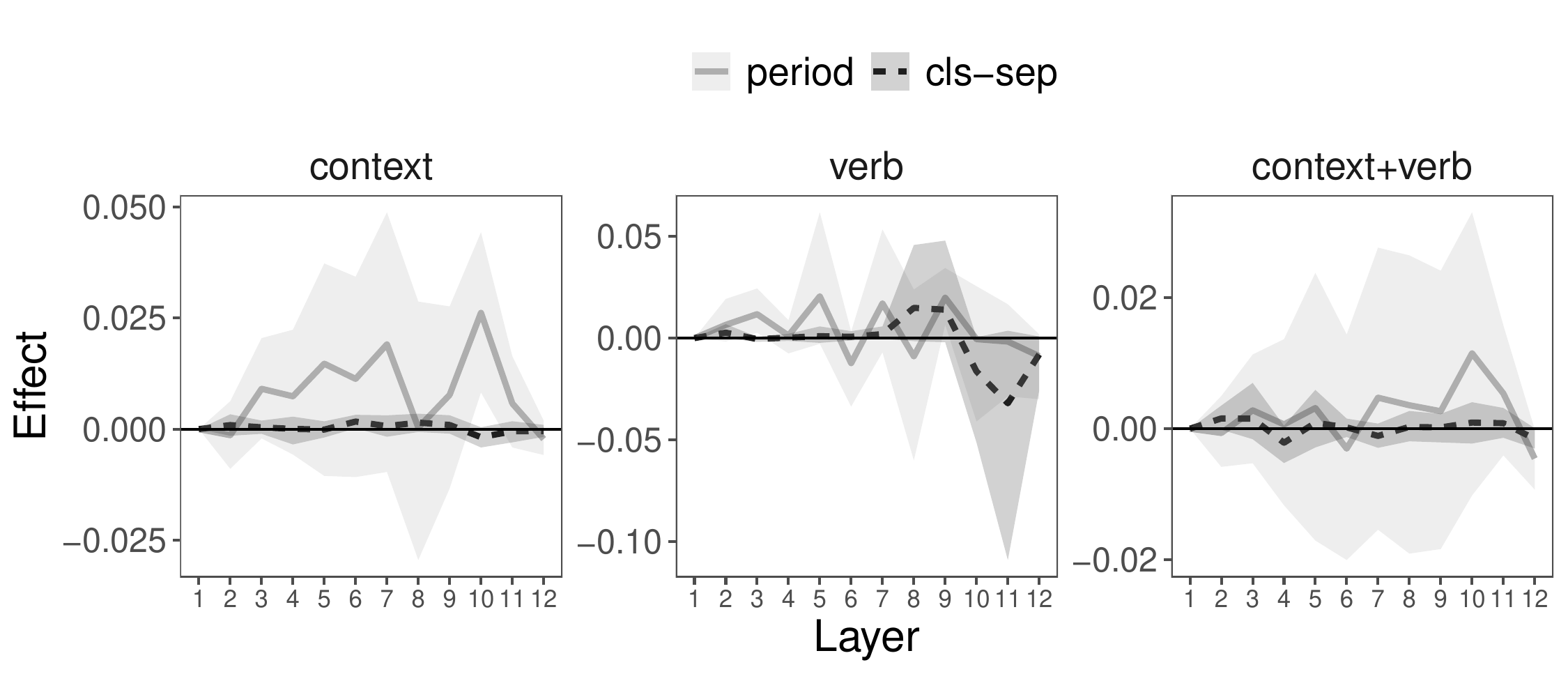}
\caption{Effects of swapping intermediate representations of periods or special tokens (i.e. \texttt{[CLS]} and \texttt{[SEP]} tokens). Error ribbons show bootstrapped 95\% confidence intervals across sentence pairs.}
\label{fig:period}
\end{center}
\end{figure*}
\begin{figure*}[th!]
\begin{center}
\includegraphics[width=0.99\linewidth]{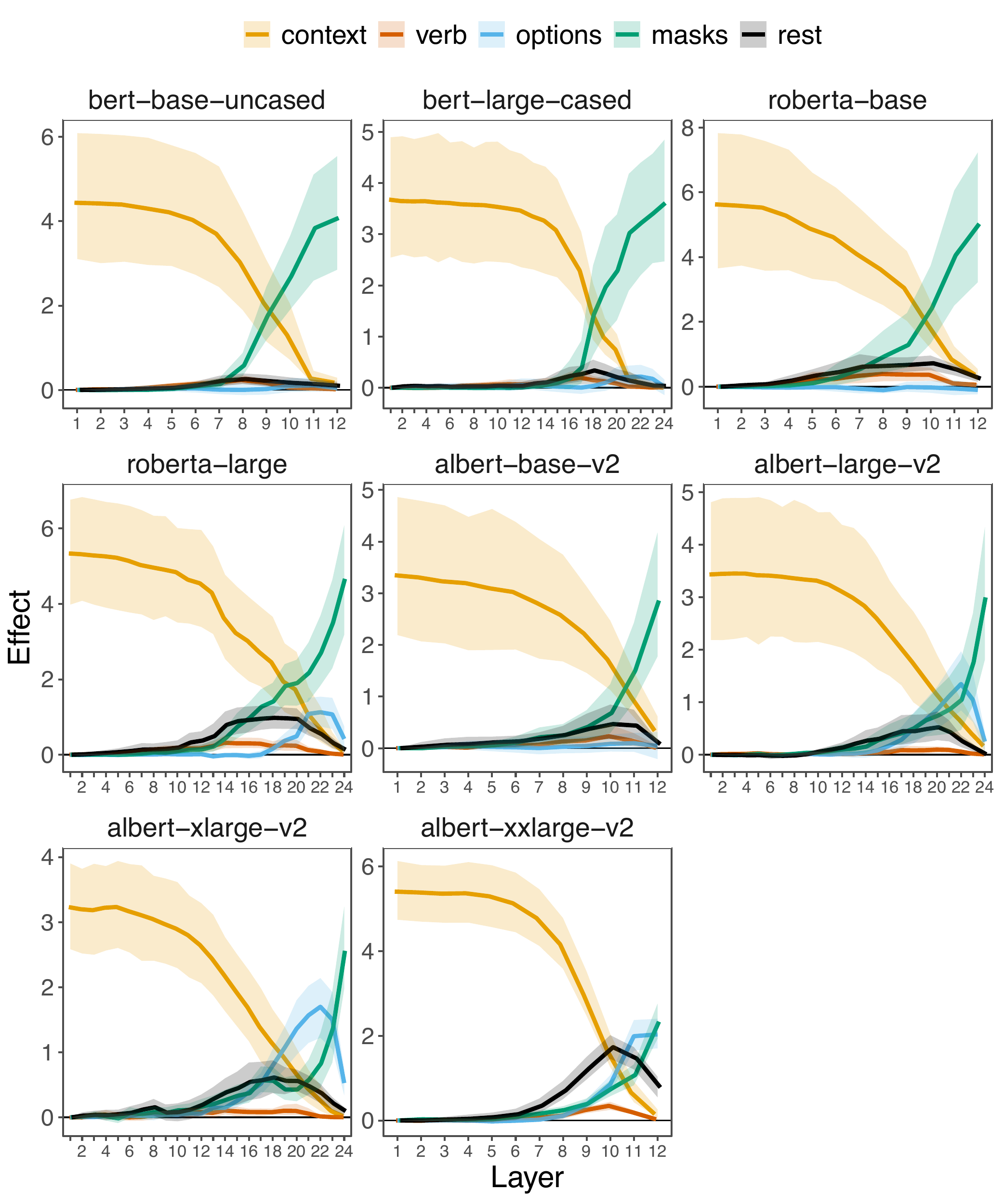}
\caption{Effect of swapping each layer for models with different types and sizes. The flow of information from the context to the mask is conserved in all models. In addition, larger models (roberta-large, albert-large-v2, albert-xlarge-v2, and albert-xxlarge-v2) have the context information flow that goes into the options and the rest.  Error ribbons show bootstrapped 95\% confidence intervals across sentence pairs.}
\label{fig:models}
\end{center}
\end{figure*}

\begin{figure*}[t!]
\begin{center}
\includegraphics[width=0.5\linewidth]{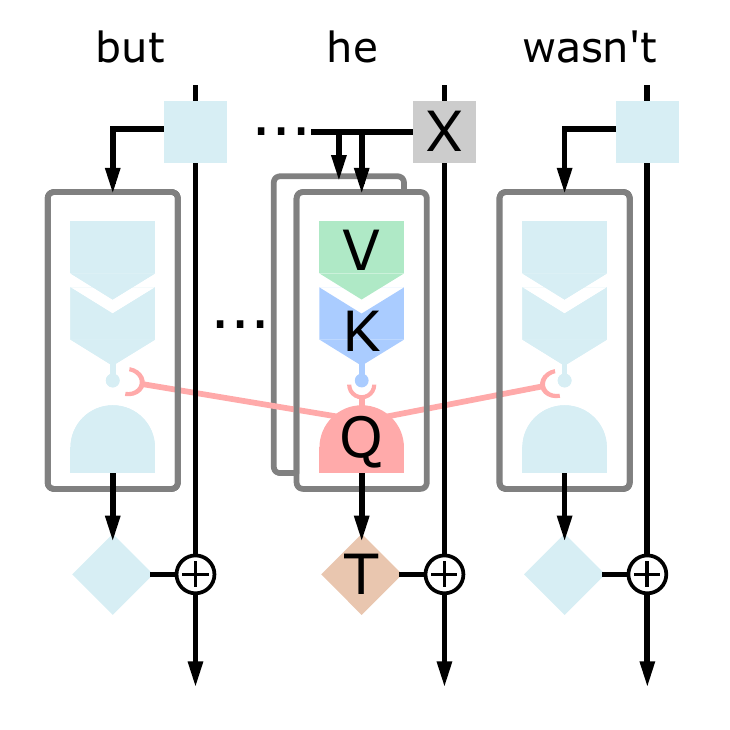}
\caption{A schematic showing how different components interact with each other in the attention mechanism. X, V, K, Q, and T refer to the layer embedding, value, key, query, transformation vectors respectively.  This schematic is used in \autoref{fig:head_context_schematic}}
\label{fig:schematic}
\end{center}
\end{figure*}

\begin{figure*}[t!]
\begin{center}
\includegraphics[width=0.66\linewidth]{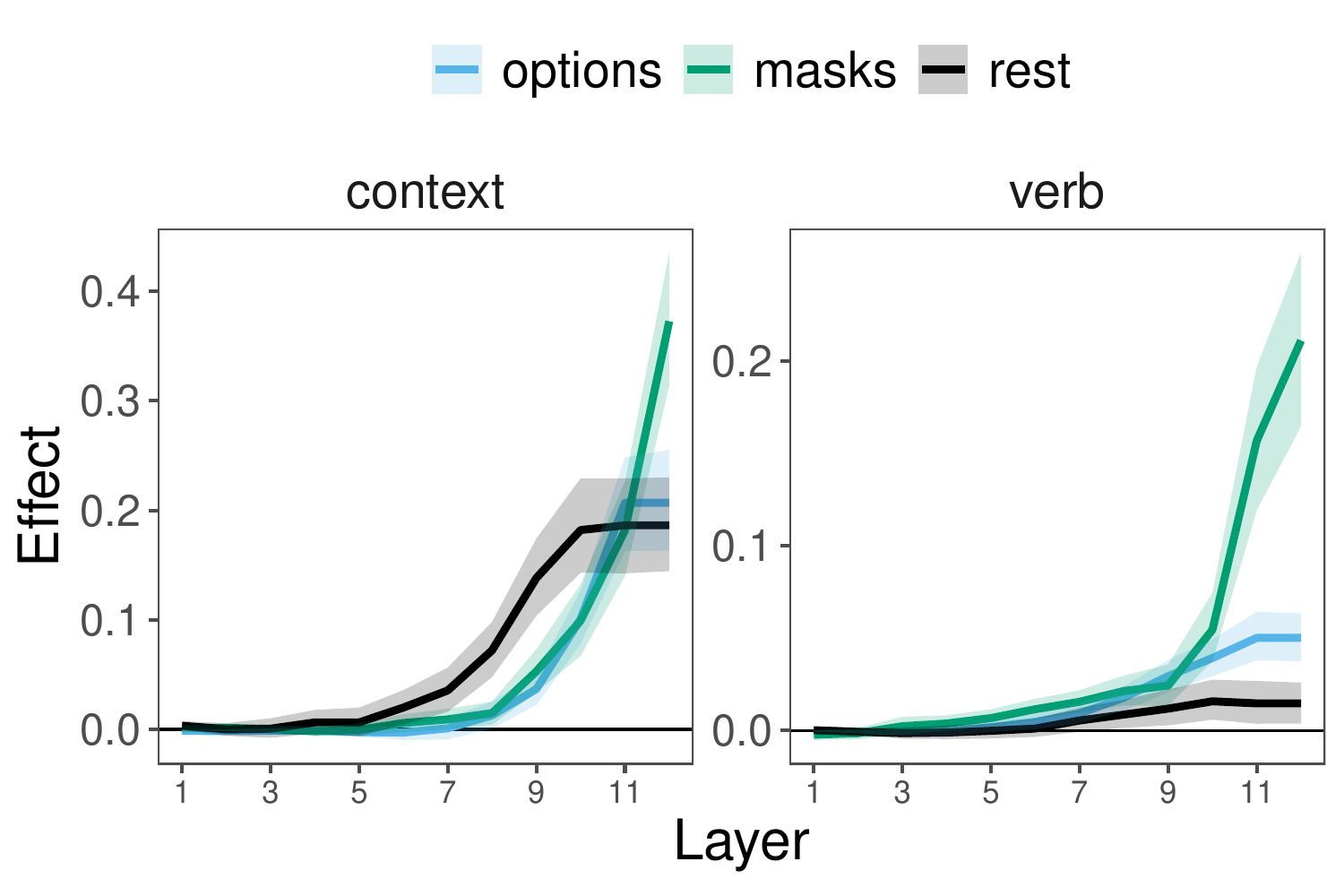}
\caption{Effects of cumulative intervention on the transformation vectors (the final layer of the attention mechanism prior to being added back to the residual stream, see \autoref{fig:schematic}). For each layer $i$, transformation vectors at layer $i$ and all the preceding layers (at layer 0, 1, ..., $i-1$) are interchanged. The cumulative effects largely reproduce the layerwise intervention \autoref{fig:layer}, which suggests the emerging context information that flows into the options, mask and rest is mediated by transformation. Error ribbons show bootstrapped 95\% confidence intervals across sentence pairs.}
\label{fig:zrep}
\end{center}
\end{figure*}

\begin{figure*}[t!]
\begin{center}
\includegraphics[width=0.99\linewidth]{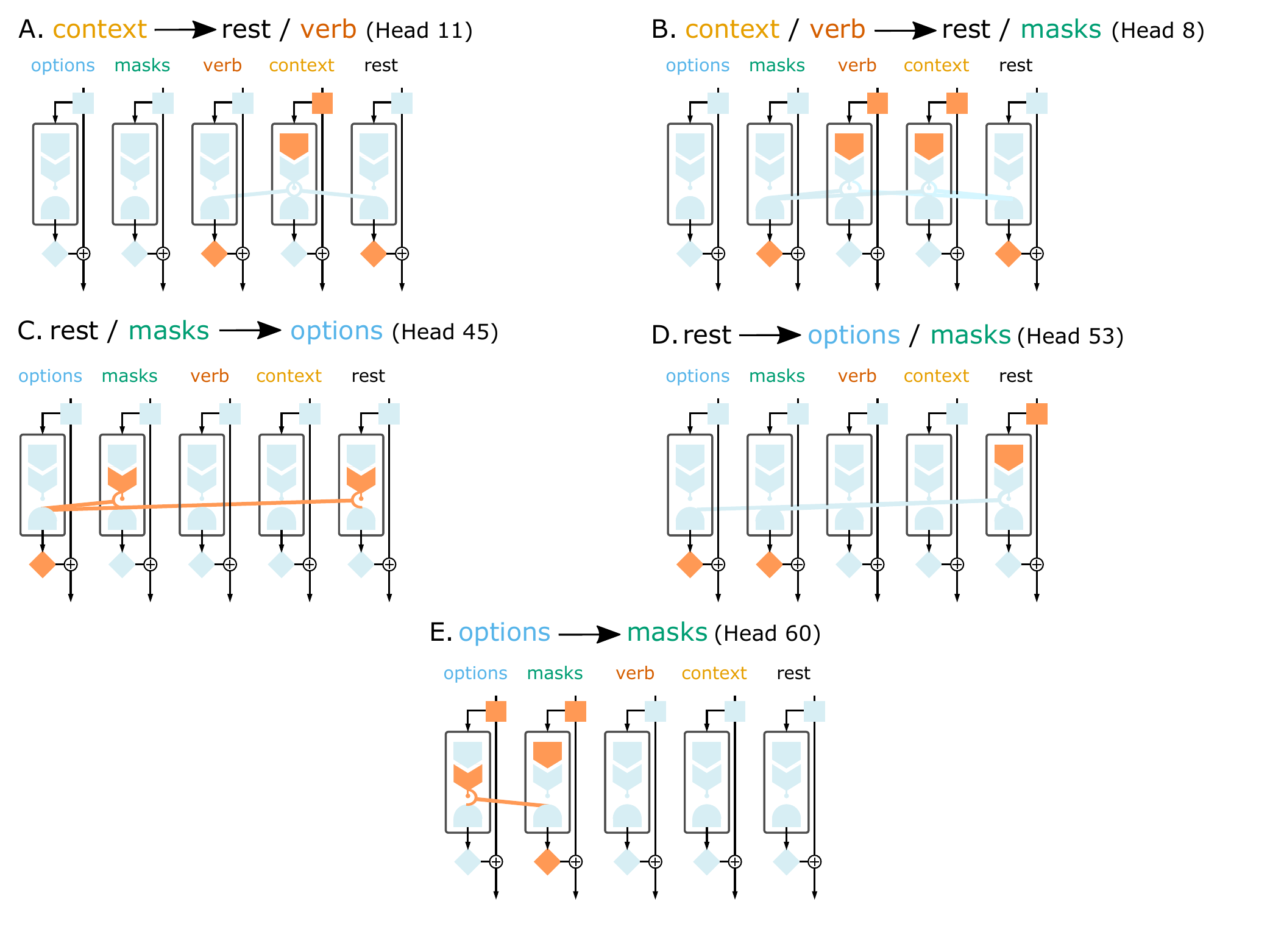}
\caption{Schematic illustration of \autoref{fig:head}. Shapes highlighted in orange are found to be implicated in the circuit.}
\label{fig:head_context_schematic}
\end{center}
\end{figure*}

\begin{figure*}[t!]
\begin{center}
\includegraphics[width=0.99\linewidth]{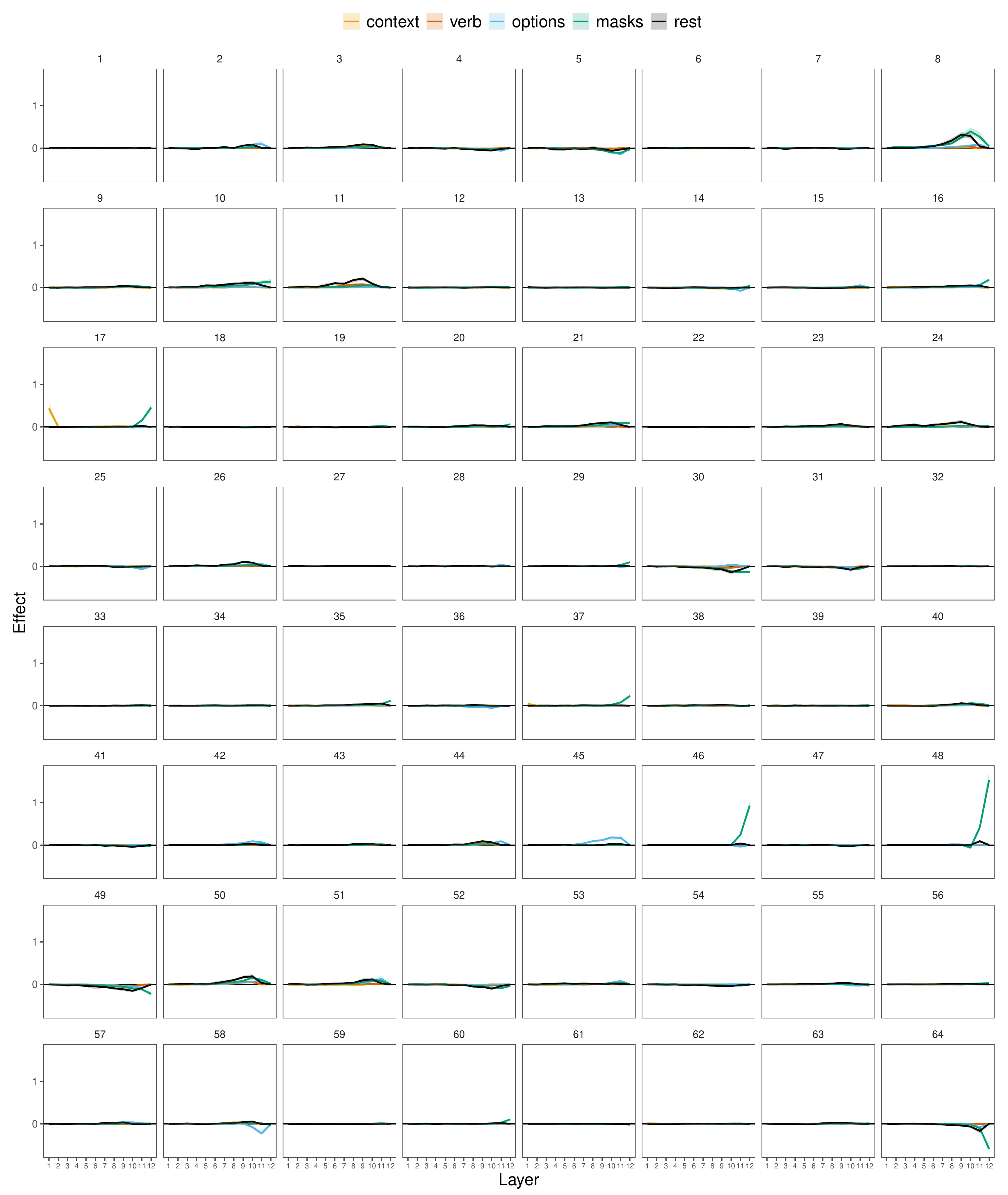}
\caption{Effects of swapping full transformation representations for each head, in the context only condition. Error ribbons show bootstrapped 95\% confidence intervals across sentence pairs.}
\label{fig:zrep_all_context}
\end{center}
\end{figure*}

\begin{figure*}[t!]
\begin{center}
\includegraphics[width=0.99\linewidth]{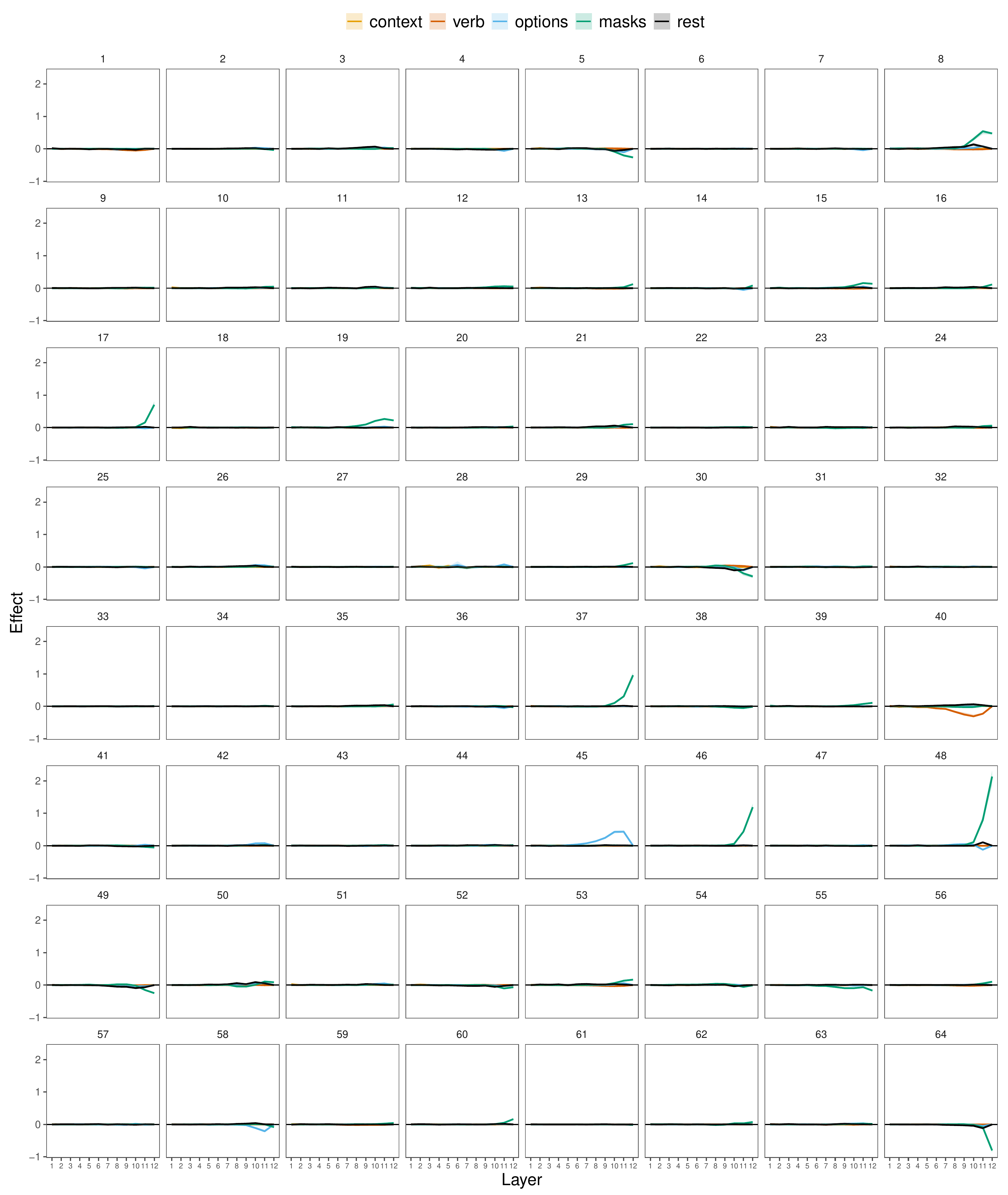}
\caption{Effects of swapping full transformation representations for each head, in the syntax only condition. Error ribbons show bootstrapped 95\% confidence intervals across sentence pairs.}
\label{fig:zrep_all_verb}
\end{center}
\end{figure*}

\begin{figure*}[t!]
\begin{center}
\includegraphics[width=0.99\linewidth]{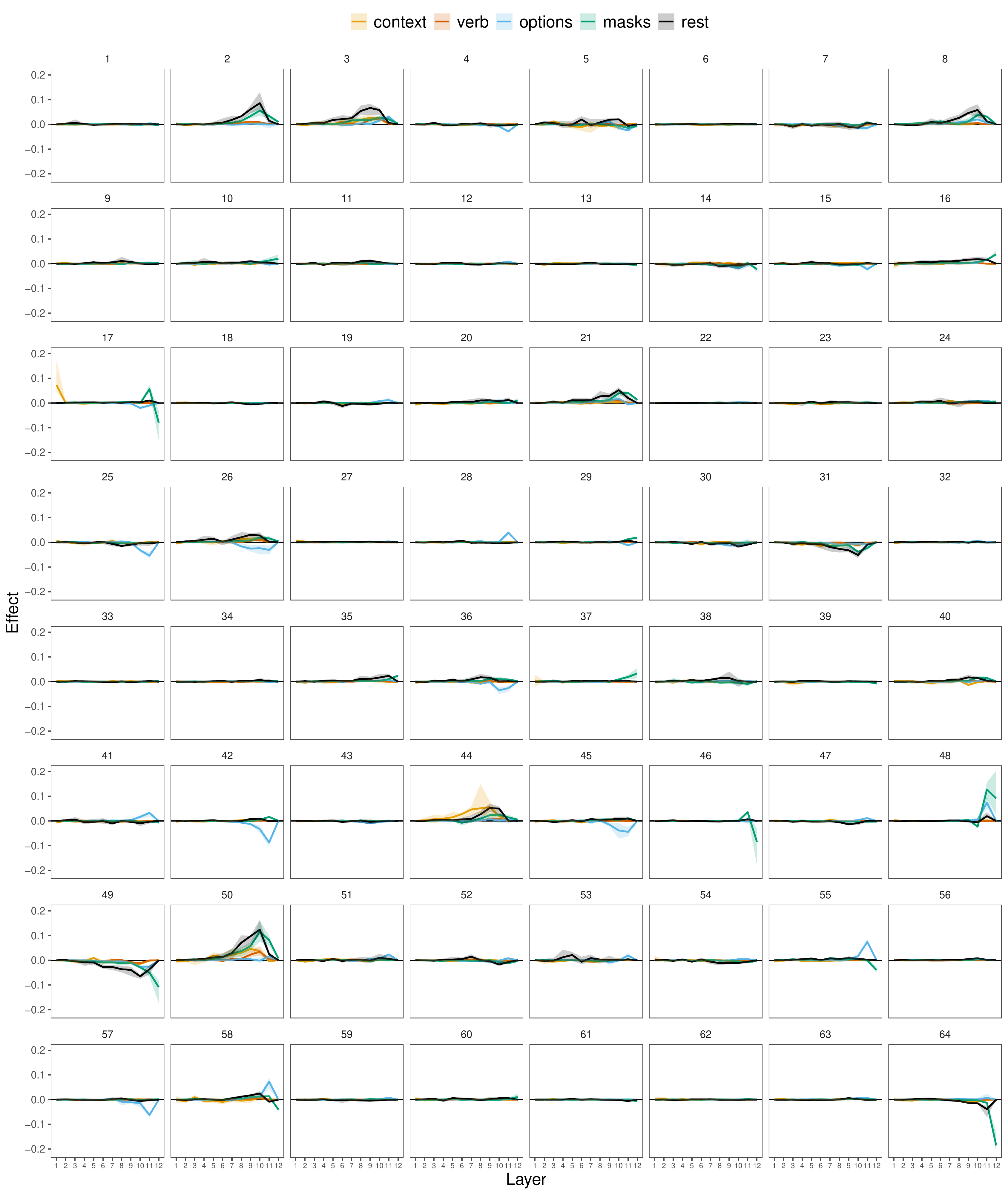}
\caption{Effects of swapping query representations for the context condition. Error ribbons show bootstrapped 95\% confidence intervals across sentence pairs.}
\label{fig:head_all_query}
\end{center}
\end{figure*}

\begin{figure*}[t!]
\begin{center}
\includegraphics[width=0.99\linewidth]{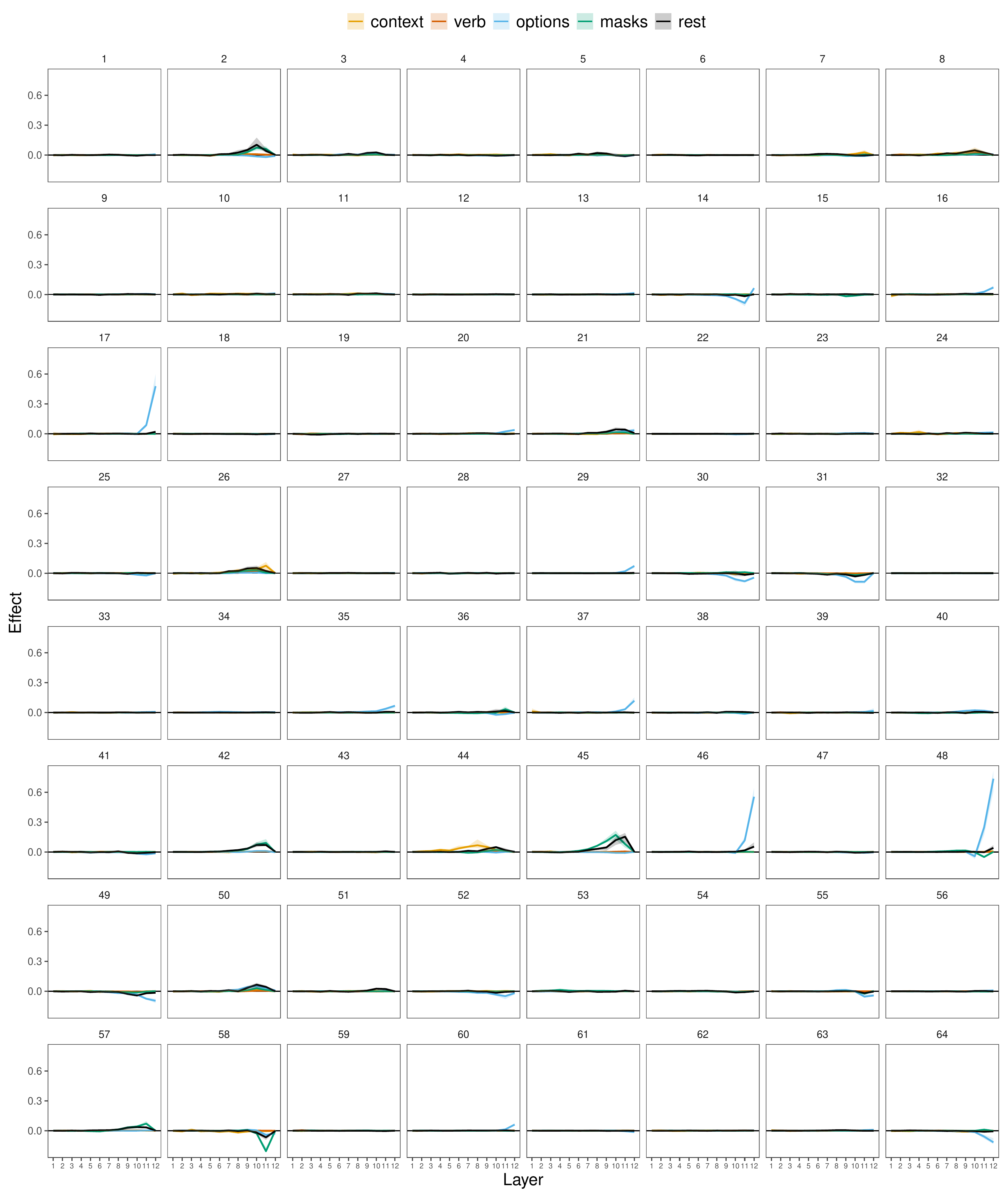}
\caption{Effects of swapping key representations for the context condition. Error ribbons show bootstrapped 95\% confidence intervals across sentence pairs.}
\label{fig:head_all_key}
\end{center}
\end{figure*}

\begin{figure*}[t!]
\begin{center}
\includegraphics[width=0.99\linewidth]{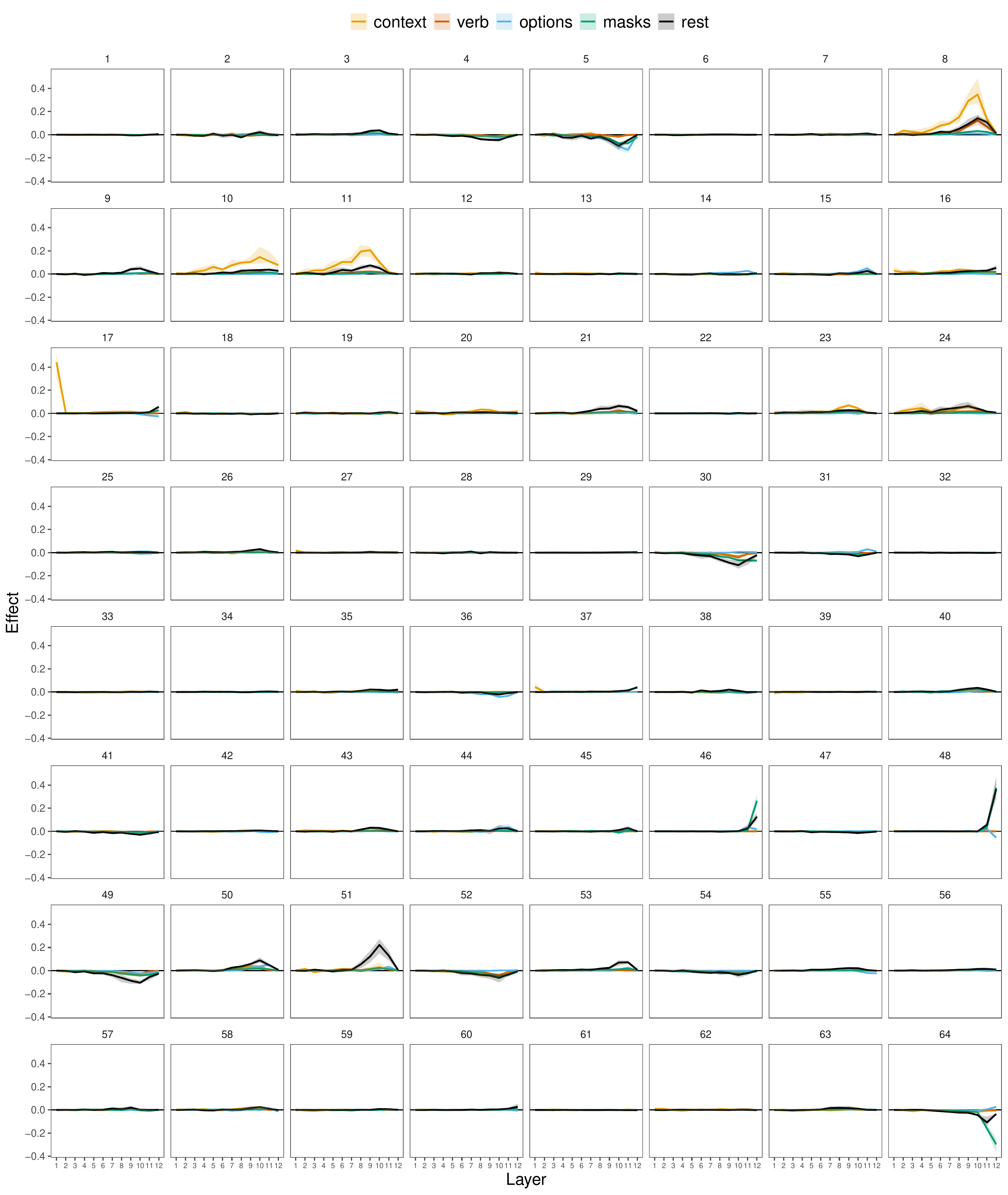}
\caption{Effects of swapping value representations for the context condition. Error ribbons show bootstrapped 95\% confidence intervals across sentence pairs.}
\label{fig:head_all_value}
\end{center}
\end{figure*}

\begin{figure*}[t!]
\begin{center}
\includegraphics[width=0.75\linewidth]{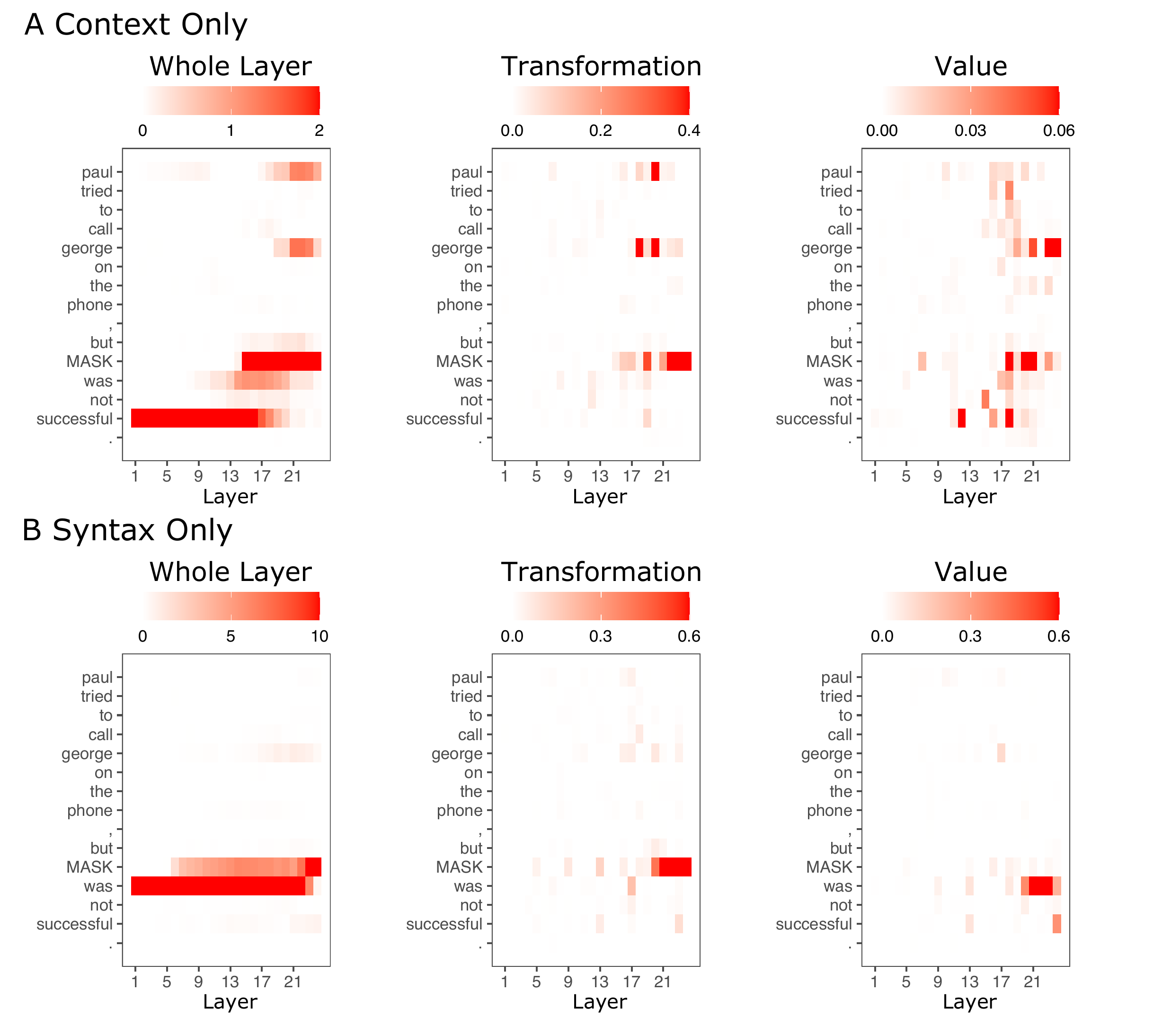}
\vspace{-1em}
\caption{In order to test the robustness of our identified circuit, we also performed the head-wise interchange intervention analysis for \texttt{roberta-large}. It was the best performing model among non-ALBERT models and showed similar layer-wise pattern in the layer-wise analysis. Here, we show the effect of causal interventions at different sites in an example sentence. Within the context (A) and verb (B) conditions, the head at each layer with the maximum $t$-value is selected.}
\label{fig:roberta}
\end{center}
\end{figure*}

\begin{figure*}[t!]
\begin{center}
\includegraphics[width=0.8\linewidth]{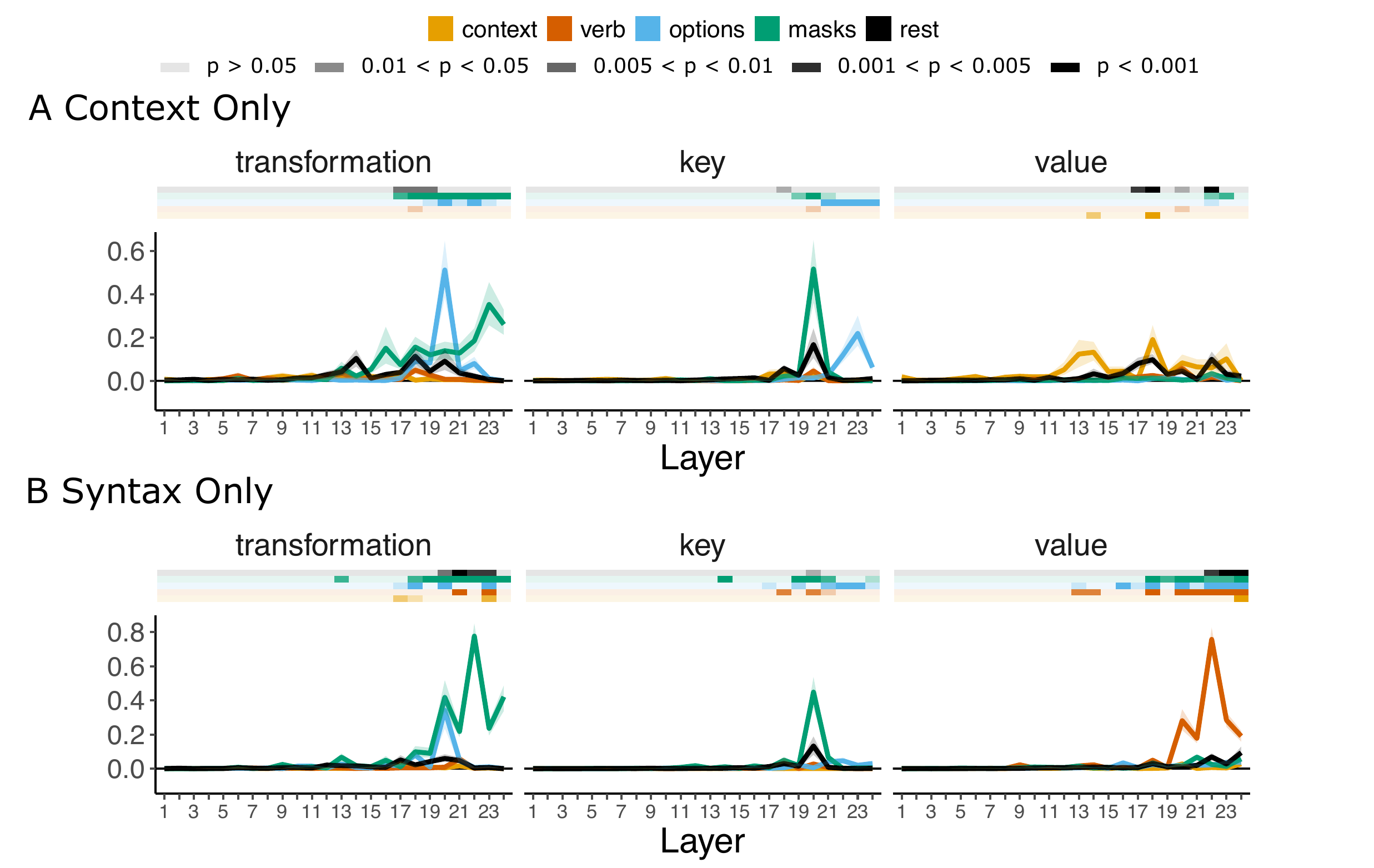}
\caption{Head-wise causal interventions aggregated across sentences for different sites and layers in RoBERTa in the context (A) and verb (B) conditions. The head with the maximum $t$-value is selected for each representation type (transformation/key/value), position (context/verb/options/masks/rest), and layer. Lines show the mean effect of the selected head, and error ribbons show bootstrapped 95\% confidence intervals.  Bars above line plots regions of statistical significance for each position.}
\label{fig:roberta_all}
\end{center}
\end{figure*}
\end{document}